\documentclass[11pt,letterpaper]{article}
\usepackage[margin=1in]{geometry}
\usepackage[T1]{fontenc}
\usepackage{lmodern}
\usepackage[round,authoryear]{natbib}

\usepackage{amsmath,amssymb,amsthm}
\usepackage{graphicx}
\usepackage{tikz}
\usetikzlibrary{arrows.meta}
\usepackage{booktabs}
\usepackage{longtable}
\usepackage{array}
\usepackage{placeins}
\usepackage{xcolor}
\usepackage{microtype}
\usepackage{xurl}
\usepackage{hyperref}

\hypersetup{
  pdftitle={Readout Scaling Shapes Sparse Feature Recovery},
  pdfauthor={Brian Xiao},
  colorlinks=true,
  linkcolor=blue!55!black,
  citecolor=blue!55!black,
  urlcolor=blue!55!black
}
\graphicspath{{./}}
\newcommand{\R}{\mathbb{R}}
\newcommand{\ones}{\mathbf{1}}

\newcommand{\rcdssae}{\textsc{RC-DS-SAE}}
\newcommand{\topk}{\operatorname{TopK}}
\theoremstyle{remark}
\newtheorem{remark}{Remark}

\title{Readout Scaling Shapes Sparse Feature Recovery}
\author{
  Brian Xiao\\
  University of Waterloo\\
  \texttt{b8xiao@uwaterloo.ca}
  \and
  Sarah Wilson\\
  Columbia University\\
  \texttt{sw4104@columbia.edu}
}
\date{}

\begin{document}
\maketitle

\begin{abstract}
\noindent Readout scaling changes what a sparse autoencoder (SAE) learns, but the benefit of
removing it depends on the data. We use readout-canonicalized decision-space (RC-DS)
coordinates to remove singular-value gains while preserving retained decision information.
At a fixed readout condition number, changing the generating distribution reverses which
training coordinates recover known factors best. In our controlled coordinate-matching
study, recovery is highest when training restores the space in which the factors were
generated. When factors are generated before readout scaling, RC-DS increases mean sparse-factor recovery from 0.506 to 0.913 at condition number 100, compared with centered-logit training under the same budget. An exact wrapper reproduces RC-DS training to numerical
precision, isolating training geometry rather than additional information. In neural
policies, RC-DS improves mean logit reconstruction under the tested budgets. On three fixed
language-model checkpoints, RC-DS has higher mean gain-free reconstruction in all 18 tested
task--sparsity settings, while centered-logit SAEs reconstruct output logits better in 17. Small-label-head controls show that longer
training reduces the reconstruction gap from 0.0198 to 0.0016. Together, these results show
how readout scaling, the data distribution, and training shape sparse learning. RC-DS provides
an information-preserving method for studying and controlling the readout's contribution.
Separate label-detection and steering tests distinguish these reconstruction effects from
task-specific feature usefulness.
\end{abstract}

\section{Introduction}

Sparse autoencoders (SAEs) decompose neural activations into a small number of active learned
features \citep{cunningham2023sae,bricken2023monosemanticity,gao2024scaling}. The usual
squared-error objective measures how well these features reconstruct the input. The choice
of input coordinates determines which reconstruction errors receive the most weight.
Two representations can contain the same information and still lead to different sparse
features. We study this effect near a neural network's linear readout.

The readout maps hidden activations to output scores, called logits, that determine action
or token probabilities. We focus on sparse features of a model's immediate decisions, so
these scores provide a direct starting point. Centering the logits over allowed outputs
removes their common offset without changing the probability distribution. An SAE trained
on these centered scores seeks reusable patterns in how the model favors some outputs
over others.

The readout scales some hidden directions more strongly than others.
If two directions are scaled by 100 and 1, equal errors before scaling contribute in the
ratio $10{,}000:1$ to squared logit error. An SAE trained on logits therefore prioritizes
the strongly amplified direction. The central question is how this readout-induced weighting
changes sparse feature recovery, and what happens when the scaling is removed.

\begin{figure}[!tb]
  \centering
  % Analytic schematic in a two-dimensional retained singular basis.
\begin{tikzpicture}[
  box/.style={draw=black!30, rounded corners=2pt, minimum height=1.9cm,
              text width=3.55cm, align=center, inner sep=7pt, font=\small},
  flow/.style={-{Stealth[length=2mm]}, thick, draw=black!60}
]
  \node[box, fill=black!3] (hidden) at (0,0)
    {\textbf{Retained state}\\[5pt]
     $q=(q_1,q_2)$\\[5pt]
     Error: $e=(e_1,e_2)$};
  \node[box, fill=orange!9] (logits) at (5.2,0)
    {\textbf{Scaled coordinates}\\[5pt]
     $\Sigma q=(100q_1,q_2)$\\[5pt]
     Loss: $10{,}000e_1^2+e_2^2$};
  \node[box, fill=blue!6] (canonical) at (10.4,0)
    {\textbf{RC-DS coordinates}\\[5pt]
     $\Sigma^{-1}\Sigma q=q$\\[5pt]
     Loss: $e_1^2+e_2^2$};
  \draw[flow] (hidden.east) -- node[above=3pt, font=\scriptsize] {readout} (logits.west);
  \draw[flow] (logits.east) -- node[above=3pt, font=\scriptsize, align=center] {remove\\gains} (canonical.west);
\end{tikzpicture}
  \caption{The same retained information, with different reconstruction weights.
  In a two-dimensional retained singular basis, the readout multiplies one direction by 100
  and the other by 1. An error $e=(e_1,e_2)$ is consequently weighted by $10{,}000:1$ in
  logit squared error. RC-DS removes these gains. The orthonormal map into output space
  is omitted because it preserves these norms.}
  \label{fig:readout-geometry}
\end{figure}
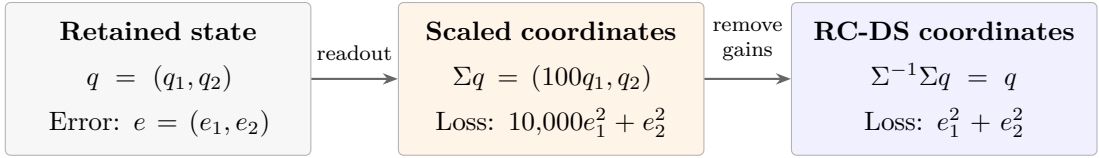

We introduce readout-canonicalized decision-space (RC-DS) coordinates. RC-DS keeps the
retained hidden directions that affect differences between output scores and removes the
readout's singular-value gains. An SAE trained on these coordinates is an \rcdssae.
At a fixed checkpoint and allowed-action set, RC-DS and retained centered logits can be
converted into each other. They carry the same decision information. Comparing them therefore
isolates a change in training geometry rather than a change in information
(Figure~\ref{fig:readout-geometry}).
Centered-logit SAEs therefore provide a controlled baseline for testing the effect of
readout scaling. Ordinary hidden-state SAEs provide a complementary comparison, testing
whether moving into readout-based coordinates improves sparse learning in the first place.

Our controlled studies show that the generating distribution changes which coordinates
recover factors best, even at the same readout condition number. Matching the training
coordinates to the generating space gives the highest recovery in the coordinate-matching
study. The condition-number sweep is a specific case: its factors are generated in the
unscaled space, and RC-DS restores that space after readout distortion. An exact wrapper
reproduces this training from scaled inputs. Readout gains and the data distribution jointly
shape recovery, so condition number alone does not determine the benefit of removing gains.

Real-model comparisons measure the corresponding reconstruction effects. We report errors
both before and after readout scaling, retain ordinary hidden-state and covariance-based
baselines, and test how longer training changes the gaps. These controls explain the
improvements through their objective and training procedure. Label detection and steering
then serve as separate application tests, summarized in Section~\ref{sec:feature-usefulness}.

\paragraph{Contributions.} We make three contributions to understanding sparse learning near
a linear readout.

\begin{enumerate}
  \item \textbf{Evidence for when removing the gains helps.} Controlled changes to the
        generating distribution reverse which coordinates recover factors better, even at the
        same condition number. A fitting-data diagnostic predicts these reversals on new
        settings within the tested synthetic family.
  \item \textbf{Controlled evidence that readout scaling changes sparse learning.}
        RC-DS comparisons preserve retained decision information while removing readout gains.
        An exact training wrapper and loss-only controls isolate how the resulting change in
        training geometry affects known-factor recovery.
  \item \textbf{A characterization of reconstruction benefits in neural policies and
        language-model readouts.} Common-metric evaluation, hidden-state and whitening
        baselines, and training-budget controls identify what improves, what is traded off,
        and how the comparison changes with training.
\end{enumerate}

\section{Related Work}

SAE research studies feature discovery, scaling, and evaluation
\citep{cunningham2023sae,bricken2023monosemanticity,gao2024scaling}.
\citet{templeton2024scaling} extend feature extraction and intervention studies to
Claude 3 Sonnet. Gated SAEs separate feature selection from magnitude estimation to reduce
shrinkage \citep{rajamanoharan2024gated}. JumpReLU SAEs improve reconstruction at a given
sparsity through their activation function and training objective
\citep{rajamanoharan2024jumprelu}. These architectural changes complement our question about
input coordinates. Our comparisons hold the SAE architecture fixed and do not establish
the same coordinate effects for every SAE variant. Benchmarking and sparse-probing
studies distinguish reconstruction, sparsity, interpretability, and usefulness
\citep{karvonen2025saebench,kantamneni2025useful}. Critiques of canonical features and random-baseline
comparisons motivate tests that isolate why a method helps \citep{leask2025canonical,korznikov2026sanity}.
We study the effect of changing readout coordinates while preserving their decision information.

Interpreting hidden states through vocabulary predictions is also the motivation for the
logit lens \citep{nostalgebraist2020logitlens} and tuned lens \citep{belrose2023tunedlens}.
The logit lens applies the model's output readout to intermediate states. The tuned lens
learns an affine translator for each layer to make these predictions more reliable.
Our study instead starts at the model's actual linear readout and changes the coordinates
used for sparse learning. It does not learn layerwise translators or add decision information.

Input scaling and encoder capacity are established concerns in sparse learning.
\citet{saraswatula2025whitening} study PCA whitening for modern SAEs and report
interpretability gains alongside reconstruction trade-offs. \citet{oneill2024amortisation}
analyze limitations of the usual sparse encoder and compare more expressive inference methods.
Our comparison isolates the scaling introduced by a linear readout. This differs from covariance
whitening, which rescales activations using their observed variances and correlations. We compare
both approaches and include a control that reproduces RC-DS training exactly.

Coordinate choice also affects optimization. Adam uses coordinatewise moment estimates
\citep{kingma2015adam}, and its behavior can change under rotations of the parameter space
\citep{zhang2025adamrotations}. This is relevant when independently trained hidden-state and
full-rank RC-DS SAEs differ despite having transportable solutions. Natural-gradient methods
address parameterization through a Fisher-based optimization metric \citep{martens2020natural}.
RC-DS is not a natural-gradient optimizer. It removes gains specified by the readout, and
ordinary squared-error training in the resulting coordinates changes the reconstruction
objective relative to logit training. Our loss-only and exact-wrapper controls examine
both reconstruction weighting and training parameterization. They do not isolate Adam
as the sole source of differences.

Representation comparisons also depend on the transformations under which a claim is intended
to hold \citep{kornblith2019similarity}. Orthogonal Procrustes alignment removes rotations without
changing Euclidean geometry \citep{schonemann1966procrustes}. RC-DS additionally removes readout
singular gains. This is not covariance whitening. The policy setting supplies exact maps from
hidden perturbations to legal logits, with OpenSpiel providing the game interface
\citep{lanctot2019openspiel}. These maps let us test geometry separately from immediate decision
information.

Our prediction test asks a different question from whether one preprocessing method improves
average benchmark scores. Geometry has also been used to predict layerwise SAE scaling
\citep{zaher2026geometric}, and statistics of trained features can predict steering side effects
\citep{duan2026preintervention}. We instead forecast the direction and size of a change between
information-equivalent readout coordinates before new SAEs are trained. The target is a paired
recovery difference, not a layer's scaling exponent or a trained feature's intervention effect.
We test semantic usefulness separately with label probes and causal interventions.

\section{RC-DS: Removing Scaling While Preserving Information}
\label{sec:rcds-method}

\subsection{Removing readout scaling}

Let $h(x)\in\R^d$ be a hidden representation and $z(x)=Wh(x)+b$ its logits for $A$ actions.
At state $x$, only the actions in $L(x)$ are allowed. The matrix $S_L$ selects those actions,
and $C_L=I-|L|^{-1}\ones\ones^\top$ subtracts their mean score. We study $W_c=C_LS_LW$
because adding the same value to all allowed logits does not change their softmax probabilities.
Write $\ell(x)=C_LS_Lz(x)=W_ch(x)+c$, with $c=C_LS_Lb$, for these centered allowed-action
logits.
For text classification, the model chooses among a fixed set of labels, all available for
every example. The equations below hold for a fixed model checkpoint and allowed-action set,
also called a \emph{mask}.

Use a singular value decomposition (SVD) $W_c^{(\rho)}=U\Sigma V^\top$, keeping the $\rho$
singular values above a declared numerical tolerance. These values, stored in $\Sigma$, are
the readout's scale factors, or \emph{gains}. RC-DS removes them. With $\mu$ the mean hidden
state on the SAE training data, define
\begin{equation}
  r_0(x)=UV^\top(h(x)-\mu).
  \label{eq:canonical}
\end{equation}
We keep the mask with each representation. The map $S_L^\top r_0$ places allowed-action
entries back into the full action vector. To compare no scaling, partial scaling, and full
scaling on the same retained subspace, use
\begin{equation}
  r_\alpha(x)=U\Sigma^\alpha V^\top(h(x)-\mu),
  \qquad \alpha\in\{0,\tfrac12,1\}.
  \label{eq:gain-family}
\end{equation}
Here $\alpha=0$ gives RC-DS, $\alpha=\tfrac12$ takes square roots of the gains, and
$\alpha=1$ keeps them. The last case is the centered-head response to $h-\mu$.
Unlike covariance whitening, this construction normalizes the readout, not the activation
distribution.

\begin{samepage}
\begin{remark}[Fixed-mask readout equivalence]
\label{rem:readout-equivalence}
Let $\ell^{(\rho)}=W_c^{(\rho)}h+c$ be the retained centered logits. On the retained subspace,
\begin{align}
  \ell^{(\rho)}&=U\Sigma U^\top r_0+W_c^{(\rho)}\mu+c,
  \label{eq:forward-equivalence}\\
  r_0&=U\Sigma^{-1}U^\top
        (\ell^{(\rho)}-W_c^{(\rho)}\mu-c).
  \label{eq:inverse-equivalence}
\end{align}
Thus, RC-DS and retained centered logits contain the same immediate decision information.
\end{remark}

\begin{proof}
Substitute $r_0=UV^\top(h-\mu)$ and use $U^\top U=I$. The forward map becomes
$W_c^{(\rho)}(h-\mu)$ plus its offset. Multiplying by $U\Sigma^{-1}U^\top$ gives the inverse.
\end{proof}
\end{samepage}

Within the retained directions, RC-DS and centered logits can be converted into each other
without losing information. They can express the same changes to output scores, but their
different scaling can affect SAE training. RC-DS specifies the input coordinates, not a unique
set of learned SAE features. Appendix~\ref{app:formal}
explains the effect of discarding small singular values and changing the hidden-state basis.

Operationally, the readout factors are fixed during SAE training. Each state is encoded using
its legal mask, and a reconstructed RC-DS vector is mapped back with
Equation~\ref{eq:forward-equivalence} before scoring. The fitting mean and actor bias are restored
as offsets, not learned again by the comparison. A mask with only one legal action has no
relative-logit degree of freedom. Different masks can have different retained ranks. We record
their spectra separately and check that mapping coordinates to logits and back recovers the
starting values within numerical tolerance.

\subsection{How scaling changes the reconstruction loss}

The condition number $\kappa=\sigma_1/\sigma_\rho$ is the ratio of the largest to smallest
retained gain. A value of one means equal scaling, and larger values mean more unequal scaling.
Write $q=V^\top(h-\mu)$ for the hidden state in the retained singular-vector basis. These
coordinates are \emph{gain-free}: the readout's scaling has not been applied. For reconstruction
error $e$ in these coordinates,
\begin{equation}
  \|U\Sigma e\|_2^2=\sum_{j=1}^{\rho}\sigma_j^2 e_j^2,
  \qquad \|Ue\|_2^2=\sum_{j=1}^{\rho}e_j^2.
  \label{eq:residual-metrics}
\end{equation}
Equation~\ref{eq:residual-metrics} gives the weighting in the introduction's two-direction
example. Removing the gains gives low-gain directions more weight relative to high-gain ones.
That can improve their reconstruction while increasing total logit error. The outcome also
depends on how much the data vary along each direction, where the factors point, how many
features may be active, and how the SAE is trained. A weakly scaled direction with little data
variation may have little effect on either error measure. Condition number also depends on
the chosen hidden-state coordinates and can change if those coordinates are rescaled.

If $\rho=d$, RC-DS keeps the whole hidden representation and preserves its Euclidean lengths
and angles. After expressing RC-DS in a hidden-width orthonormal basis, the two representations
differ only by a rotation or reflection. Transporting an SAE's encoder, decoder, and bias
through that transformation preserves its activations, sparsity, and Euclidean reconstruction
error. This gives equivalent models. Separately trained models follow their own optimization
paths because Adam's coordinatewise updates depend on the basis \citep{kingma2015adam}. Ordinary hidden-state
SAEs are therefore essential controls in this case.

\subsection{Training and evaluation}

An SAE learns a dictionary of feature directions, also called decoder atoms. Its encoder
assigns nonnegative weights to at most $K$ directions per input, and its decoder adds them
to reconstruct that input. We use unit-length decoder atoms and a Top-$K$ encoder
\citep{gao2024scaling},
\begin{equation}
  f(s)=\topk_K(\operatorname{ReLU}(W_e(s-b_d)+b_e)),
  \qquad \hat s=Df(s)+b_d.
\end{equation}
Our \emph{exact wrapper} tests whether a result can be reproduced by undoing the coordinate
change. It removes the scaling before the encoder, restores it after the decoder, and removes
it again when measuring error. For scaled inputs $y=Tr$, where
$T=U\Sigma^\alpha U^\top$ is invertible on the retained subspace, define
\begin{equation}
  \hat y=T F_\theta(T^{-1}y),
  \qquad \mathcal L=\|T^{-1}(y-\hat y)\|_2^2=\|r-F_\theta(r)\|_2^2.
  \label{eq:wrapper}
\end{equation}
Here $F_\theta$ is the SAE trained on RC-DS inputs. With the same parameters and optimizer
state, the two procedures give the same activations, gradients, and updates. Reweighting the
loss alone does not guarantee this result because the encoder still receives scaled inputs.

\emph{Reconstruction} measures how closely an SAE reproduces a specified target.
We score every method in the same two spaces. In output space, we convert predictions to
centered logits and compute explained variance (EV). Let $\ell_i=\ell(x_i)$ be the target,
$\hat\ell_i$ its reconstruction, and $\bar\ell$ the mean target over evaluation examples:
\begin{equation}
  \mathrm{EV}=1-\frac{\sum_i\|\ell_i-\hat\ell_i\|_2^2}
                       {\sum_i\|\ell_i-\bar\ell\|_2^2}.
  \label{eq:explained-variance}
\end{equation}
Higher EV means better reconstruction, with one indicating zero error. Lower mean squared
error (MSE) likewise means better reconstruction. For \emph{gain-free EV}, we apply the same
formula after undoing the readout's scaling in predictions and targets. This measures fidelity
to the retained hidden representation. It changes the weight of each error, not the predictions
being evaluated.
The full-vocabulary reconstruction analysis reports both as secondary outcomes of a study
whose primary outcome was semantic label detection.

\emph{Factor recovery} connects reconstruction to learned features by comparing learned
directions with the known factors in synthetic
data. We map learned directions back to the generating coordinates, find the best one-to-one
match, and average their cosine similarities. Higher scores mean closer alignment. We retain
the sign because opposite directions are distinct for nonnegative codes. The score measures
directional alignment. Real-model studies instead measure reconstruction, label detection,
and steering because their generating factors are unknown.

All preprocessing statistics use SAE training data only.
We report training seeds and active-feature counts, and distinguish reused development data
from new evaluation data. Reported measurements use at most four significant figures.
Differences are computed before rounding, and saved results retain full precision. Counts,
seeds, and configured training settings are reported exactly.
Section~\ref{sec:feature-usefulness} defines the separate
label-detection and steering evaluations.

\begin{samepage}
\subsection{Coordinate baselines}

Centered logits are the main baseline because they contain the same retained decision
information as RC-DS, with the gains left in place. A hidden-state SAE is a second essential
baseline. When the readout has a nullspace, it also sees directions that cannot affect output
scores. At full column rank, it contains the same information as RC-DS and differs only by
an orthogonal coordinate change. This is the case for the three full-vocabulary readouts.
The same number of dictionary atoms does not imply the same parameter count when input
dimensions differ, as in the smaller label-head comparisons.
\par
\end{samepage}

For the full-vocabulary study, we omit the orthonormal map $U$ and use one input coordinate
per retained singular direction. We call $q$ \emph{compact RC-DS} and $\Sigma q$
\emph{compact logits}. Both exclude the fixed offsets, which are restored when scoring
output logits.

\begin{table}[!ht]
  \centering
  \small
  \caption{Baselines and what they test. Reconstruction comparisons use common logit and
  gain-free metrics. The $q$ coordinates avoid storing a vocabulary-wide input in the full-readout
  study. The exact wrapper is tested on synthetic data.}
  \label{tab:coordinate-controls}
  \begin{tabular}{@{}p{0.30\linewidth}p{0.65\linewidth}@{}}
    \toprule
    Input or training control & Explanation tested \\
    \midrule
    Centered logits, $U\Sigma q$ & Keep the readout's scaling. \\
    Hidden states, $h-\mu$ & Keep all hidden directions, including those invisible to the readout. \\
    Compact logits, $\Sigma q$ & Keep the gains but omit the map $U$ into output space. \\
    Compact RC-DS, $q$ & Remove the gains and omit $U$. \\
    Coordinate standardization & Rescale each input coordinate by its standard deviation. \\
    PCA / ZCA whitening & Rescale using both variances and correlations in the data. \\
    Inverse-gain loss only & Undo scaling in the error measure, but not in the encoder input. \\
    Exact wrapper & Undo the full coordinate change and reproduce RC-DS training. \\
    \bottomrule
  \end{tabular}
\end{table}

Whitening uses the covariance of observed states, whereas RC-DS uses the readout matrix
(Table~\ref{tab:coordinate-controls}). The full-vocabulary study uses one coordinate per retained
singular direction to match input sizes and parameter counts. This preserves reconstruction
error. Compact and vocabulary-wide parameterizations can follow different Adam updates.
The exact wrapper retains the same parameters and optimizer state to reproduce training.
The appendix gives the numerical safeguards used for whitening.

The difference from whitening can be written directly. For a fixed mask, let
$\Gamma=\operatorname{Cov}(q)$ on the fitting distribution. Ignoring constant offsets,
\begin{equation}
  \operatorname{Cov}(r_0)=U\Gamma U^\top,
  \qquad
  \operatorname{Cov}(\ell^{(\rho)})=U\Sigma\Gamma\Sigma U^\top.
  \label{eq:covariance}
\end{equation}
RC-DS removes $\Sigma$ but leaves the unequal variances and correlations in $\Gamma$ intact.
Whitening normalizes the observed covariance, which combines readout gains with the distribution
of hidden states.

The baselines separate these two normalizations. The contribution of direction $j$ to logit variance is
$\sigma_j^2\Gamma_{jj}$, so the singular spectrum alone does not describe the errors that
dominate training.

\section{Why Scaling Changes Feature Recovery}
\label{sec:synthetic-mechanism}

Synthetic data isolate how readout scaling affects recovery of known features. We first change
the generating geometry while keeping the readout spectrum fixed. We then examine the
condition-number sweep as a specific case and use training controls to explain its recovery gap.

\subsection{The generating geometry changes which coordinates win}
\label{sec:coordinate-matching}

We add two of 16 unit-length random directions in eight dimensions with nonnegative weights,
then center the resulting samples to obtain $s$. At fixed $\kappa=100$, output coordinates are
$y=\Sigma^g s$, with $g\in\{0,\tfrac12,1\}$. Each value of $g$ places the same base factors
in a different coordinate system. The readout remains $\Sigma$, and the corresponding
hidden coordinates are $q=\Sigma^{g-1}s$, so $y=\Sigma q$ in every case.

Training uses $\Sigma^{-t}y$, with $t\in\{0,\tfrac12,1\}$. No removal ($t=0$) gives the
compact-logit baseline, half removal gives the intermediate coordinates, and full removal
($t=1$) gives RC-DS. Within each row, these invertible transformations preserve the information
supplied to the SAE. We map learned directions back to the base generating space before
scoring recovery. The comparison uses seeds 440--451, 2,048 samples, and 40 epochs per fit.

\begin{figure}[!ht]
  \centering
  \includegraphics[width=0.78\linewidth]{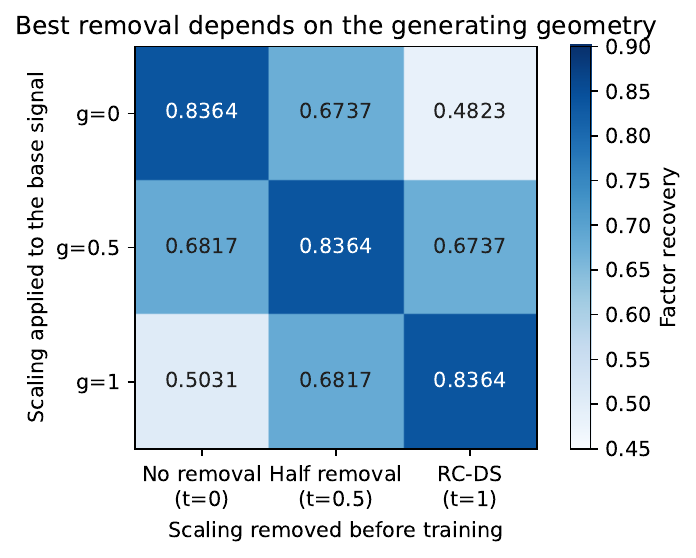}
  \caption{Changing the generating geometry changes which training coordinates recover
  factors best, with the readout condition number fixed at 100. Rows apply $\Sigma^g$ to the
  base sparse signal, and columns remove $\Sigma^t$ before training. Scores average twelve
  paired seeds after 40 epochs. The diagonal returns the same training inputs, so its equal
  scores are a construction check, not independent replications. The off-diagonal scores
  measure the effect of mismatched scaling.}
  \label{fig:synthetic-matching}
\end{figure}

The preferred training coordinates change with the generating geometry
(Figure~\ref{fig:synthetic-matching}). When $g=1$, RC-DS recovers the base space and scores
0.8364, compared with 0.5031 for the logit baseline. When $g=0$, the logit baseline already
receives the base signal and scores 0.8364, while RC-DS scores 0.4823.

Recovery is highest at $t=g$ in every row. All three diagonal means are 0.8364 because the
transforms cancel, yielding the same inputs, initialization, and minibatch schedule up to
numerical precision. The empirical result is the recovery loss away from this diagonal and
the resulting change in the preferred method. It identifies a coordinate-matching effect in
this generator, not a generally best representation.

\subsection{A specific case: factors generated before readout scaling}
\label{sec:synthetic-conditioning}

\begin{figure}[!ht]
  \centering
  \includegraphics[width=\linewidth]{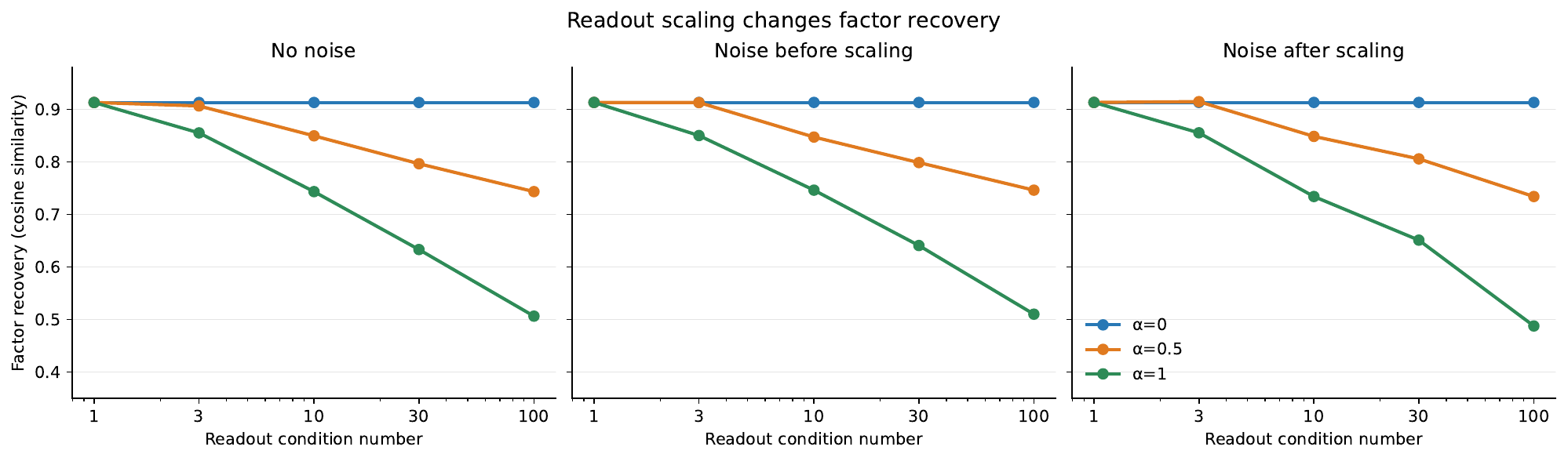}
  \caption{The case where factors are generated in the unscaled hidden space.
  Unequal readout scaling reduces mean recovery under ordinary logit training in this generator.
  Results average seeds 200--219 after 60 epochs. RC-DS ($\alpha=0$) receives the same inputs at
  every condition number. Noise added before scaling is transformed with the signal, so
  the methods still receive the same information.}
  \label{fig:synthetic-conditioning}
\end{figure}

The condition-number sweep keeps the $g=1$ construction: unit-length factors are generated
in the unscaled hidden space before the readout applies its gains. We vary the scale factors
in Equation~\ref{eq:gain-family}. The 20-seed sweep (seeds 200--219)
uses 4,096 samples and 60 SAE epochs at each $\kappa\in\{1,3,10,30,100\}$. With full scaling
and no noise, mean
recovery falls from $0.9134$ to $0.8554,0.7435,0.6332,0.5063$ as $\kappa$ increases.
RC-DS removes the scaling, returning the same inputs and recovery 0.9134 at every condition
number (Figure~\ref{fig:synthetic-conditioning}).

We also add noise before or after scaling. When noise is added before scaling, the same
invertible transform acts on the signal and the noise. The methods therefore still receive
the same information. With this pre-gain noise at $\kappa=100$, RC-DS improves recovery by
0.4034, with a paired seed-bootstrap interval of $[0.3627,0.4445]$, despite identical
information in both inputs. The no-noise difference is 0.4070
(Table~\ref{tab:app-noise}).

\subsection{The recovery gap comes from training geometry}

\begin{figure}[!ht]
  \centering
  \includegraphics[width=0.78\linewidth]{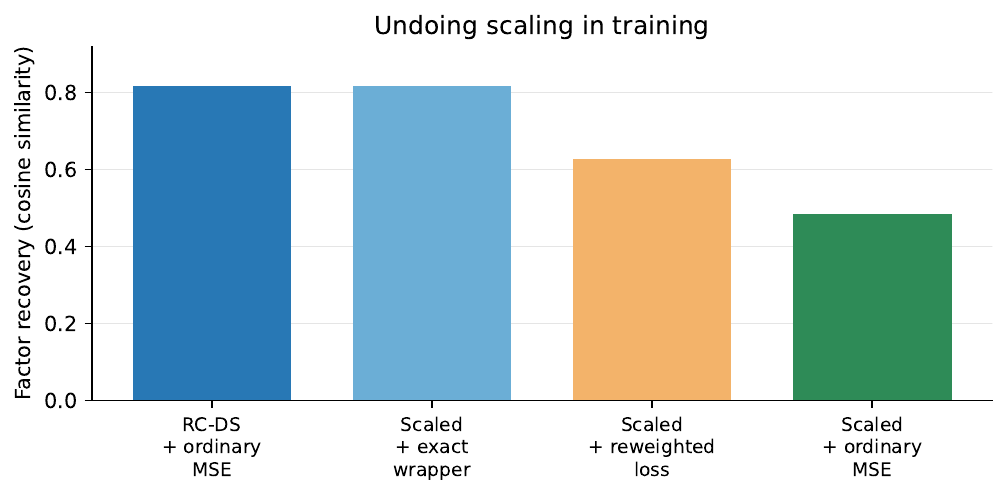}
  \caption{Training controls for the case with factors generated before readout scaling.
  RC-DS and the exact wrapper both score 0.8176, versus 0.6284 when only the loss is
  reweighted and 0.4841 with ordinary scaled-input MSE. The exact wrapper changes the input,
  output, and loss together. Means use seeds 400--411, 2,048 samples, 40 epochs, and
  $\kappa=100$.}
  \label{fig:synthetic-attribution}
\end{figure}

A separate 12-seed comparison (seeds 400--411) keeps factors in the unscaled generating space
and asks whether changing the loss alone explains the recovery gap at $\kappa=100$.
Using 2,048 samples and 40 epochs, RC-DS and the exact wrapper match in every trial, with mean
recovery 0.8176. Keeping the gains and using ordinary MSE gives 0.4841. Undoing the gains only
in the loss gives 0.6284 (Figure~\ref{fig:synthetic-attribution}). Loss reweighting therefore
recovers part, but not all, of the difference under this training procedure.
Separately, nine numerical checks spanning $\kappa=3,30,300$ compare initial and trained
activations, reconstructions, decoder directions, initial gradients, and final parameters.
Their maximum discrepancy is $1.33\times10^{-15}$, consistent with the exact equivalence in
Equation~\ref{eq:wrapper}.

\begin{samepage}
The condition-number sweep uses a different sample size and training budget from the
coordinate-matching and loss comparisons. Each study also uses separate seeds, so their
absolute scores should not be compared directly. Within each study, paired seeds
give methods the same sampled problem, and bootstrap intervals resample seeds rather than
individual activations. In the loss comparison, RC-DS improves on ordinary scaled-input MSE by
0.3335 with interval $[0.2851,0.3816]$. An untuned alternative trainer retains a positive difference
of 0.1656 with interval $[0.1377,0.2008]$.
\par
\end{samepage}

\FloatBarrier
\subsection{Correlation, sparsity, and training also change the result}

The coordinate-matching result comes from a controlled generator. To test its sensitivity, we vary
how similar the generating directions are and how many contribute to each sample.
The comparison uses two correlation settings, one, two, or four active factors, two condition
numbers, and two learning rates. The SAE's $K$ equals the number of active generating factors.
Four coordinate methods and three paired seeds give 288 fits, with encoder and decoder weights
updated independently. At condition number one, RC-DS and logit recovery agree to numerical
precision. At condition number 100, RC-DS has higher mean recovery in 10 of 12 settings.
The ordering depends on correlation, simultaneous activity, and training settings, even when
factors are generated before scaling. Appendix~\ref{app:fresh-tradeoffs} gives the full
construction and results at both learning rates.

\subsection{Predicting the direction before training}
\label{sec:observable-recovery}

We forecast the preferred coordinates using only fitting-example geometry: differences in
angular overlap and covariance conditioning,
together with the configured sparsity. The predictor does not receive true factors, their
coefficients, or evaluation examples. Unlike direction similarity, the outcome here is support
F1: how often the SAE activates the correct generating factors, after a one-to-one assignment
selected on fitting examples and held fixed for evaluation.

Six development trials select a four-term regression. After freezing it, we issue forecasts
for twelve new settings with three seeds each, then train all 288 SAE fits. The prediction
gets the sign right in all 36 pairs. MAE is 0.0526, versus 0.241 for predicting zero effect.
The MAE misses the declared 0.05 error target. Each of the twelve settings contributes three
paired seeds. ZCA has higher support F1 than RC-DS in 34 of the 36 pairs.

The useful distinction is not high versus low condition number. At the same condition number
100, with two active factors and zero construction correlation, moving the generating factors
changes an RC-DS benefit of $+0.4095$ F1 into a cost of $-0.4460$. The observable diagnostic
predicts both. The readout and the data must be considered together.
Appendix~\ref{app:observable-recovery} gives the formula, frozen development procedure,
noise and generating-coordinate shifts, all selection rules, and uncertainty.

\begin{figure}[!ht]
  \centering
  \includegraphics[width=\linewidth]{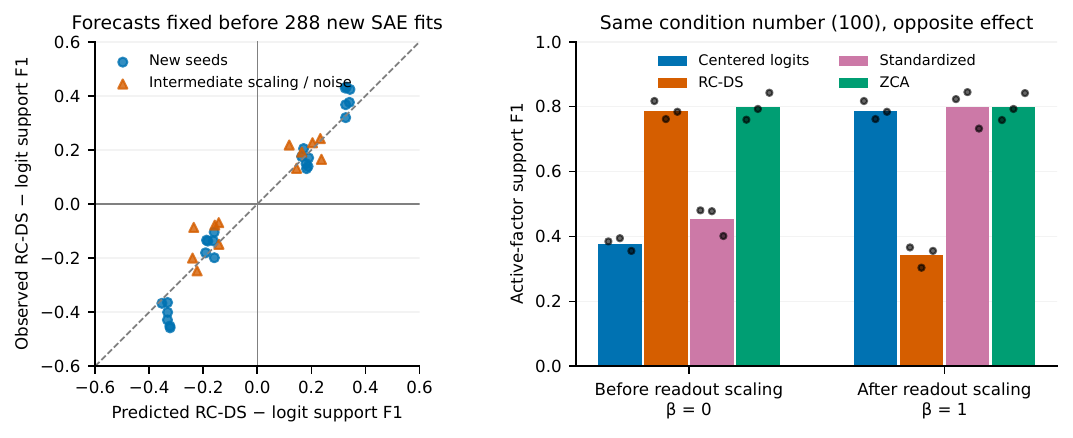}
  \caption{Fitting-data geometry predicts opposite recovery effects at the same readout
  condition number. Left: all 36 frozen forecasts, covering twelve settings and three seeds.
  Right: two matched settings with condition number 100, two active factors, and zero
  construction correlation. Bars average three seeds, shown as dots. Moving the generating
  factors reverses the RC-DS--logit ordering. Whitening is a separate, strong control.}
  \label{fig:observable-prediction}
\end{figure}

\section{Reconstruction in Neural Policies and Language Models}
\label{sec:reconstruction}

The controlled studies identify changes in known-factor recovery. In real models, the
generating factors are unknown, so we measure which parts of the representation the SAE
preserves. We first compare reconstruction in policies and full-vocabulary readouts, then
use alternative coordinates and longer training to explain the measured gaps. All differences
are computed from unrounded means. Means and differences are rounded separately for display,
so subtracting rounded table entries can differ by 0.0001 from the reported difference.
Decimal places record numerical summaries, not certainty about effects across models.

\subsection{RC-DS improves fixed-budget policy reconstruction}

In OpenSpiel bargaining, mean centered-logit EV rises from 0.8802 to 0.9144 under RC-DS
training. In seven-card Goofspiel it rises from 0.7654 to 0.8637. Both comparisons use 192-atom,
Top-$K$ 8 SAEs, trained for 15 epochs in bargaining and 125 in Goofspiel. The gain is positive
for all three bargaining policies and two of three Goofspiel policies. Goofspiel policy seed
1120 has a decrease of 0.0045. Bargaining's policies had already been inspected during
development, whereas Goofspiel uses fresh policy seeds under its retained protocol.
Appendix~\ref{app:policy-reconstruction} gives the per-policy values and fixed-budget protocols.

\subsection{Full-vocabulary readouts expose a metric trade-off}

The full-vocabulary comparison uses Qwen2.5-0.5B \citep{qwen2024qwen25}, Qwen3-0.6B-Base
\citep{qwen2025qwen3}, and SmolLM2-360M \citep{benallal2025smollm2}. We evaluate each model on
Emotion \citep{saravia2018carer}, AG News \citep{zhang2015character}, and IMDb
\citep{maas2011imdb}. The readout condition numbers are 22.6, 14.4, and 79.4,
respectively.
Each model--dataset pair has 2,400 fitting texts and 1,200 disjoint
scoring texts. Five coordinate methods use 1,024 dictionary atoms, $K\in\{8,32\}$, three
paired seeds, and two learning rates. All train for 6,000 updates, giving 540 fits.
Each method selects its rate using only its final fitting loss. Model weights remain frozen.
Appendix~\ref{app:full-readout-usefulness} gives the full data split, controls, and training
procedure, including the separate 600-text feature-selection split used for label detection.

\begin{figure}[!ht]
  \centering
  \includegraphics[width=\linewidth]{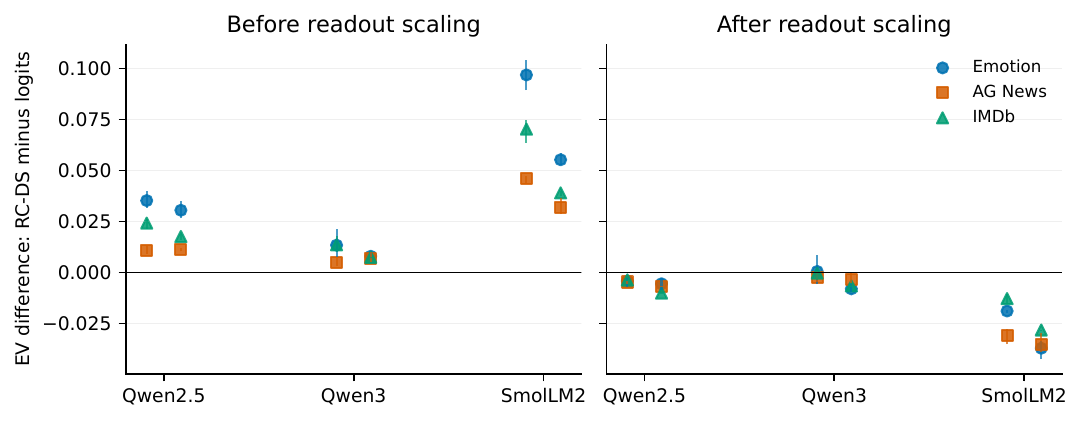}
  \caption{Changing the reconstruction target changes the preferred coordinates.
  The same fitted SAEs are scored before scaling (left) and on output logits (right).
  Positive values favor RC-DS over centered-logit training. Each point averages three paired
  SAE seeds for one dataset and sparsity level. Vertical lines span the three paired
  differences, not confidence intervals. $K=8$ is to the left and $K=32$ to the right
  of each model label. The 18 points per panel summarize the nine model--dataset pairs at two
  sparsity levels, not independent model replications.
  The right panel shows the corresponding output-score trade-off.}
  \label{fig:reconstruction-tradeoff}
\end{figure}

RC-DS has higher mean gain-free EV than centered-logit SAEs in every one of the 18 settings.
The paired difference is positive for each of the three observed seeds in each setting.
This is a descriptive count across three fixed checkpoints, three tasks, and two sparsity
levels, not 18 independent model replications or 18 significance tests. We pair SAE seeds
within each setting. Figure~\ref{fig:reconstruction-tradeoff} shows the mean and observed
seed range of each difference. Table~\ref{tab:reconstruction-summary} averages tasks and
sparsities within each seed before summarizing each checkpoint. Appendix~\ref{sec:paired-reconstruction}
retains paired ranges against every coordinate baseline, including small and negative differences.
The gains range from 0.0048 to 0.0968, with the largest differences on SmolLM2
(Figure~\ref{fig:reconstruction-tradeoff}). For example, on SmolLM2 Emotion with eight active
features, gain-free EV rises from 0.5428 to 0.6397. On Qwen2.5 Emotion at the same sparsity,
it rises from 0.4269 to 0.4622.

Centered-logit SAEs have higher mean logit EV in 17 of 18 settings, even though they have
lower gain-free EV in all 18. This is the weighting trade-off in
Equation~\ref{eq:residual-metrics}: preserving strongly amplified output directions and
preserving the unscaled representation are different objectives. Both scores were recorded
for the same fits. Reconstruction is a secondary analysis of a study with a primary
label-detection outcome, and the present emphasis was chosen after examining the results.
These full-vocabulary fits use a fixed 6,000-update budget. Section~\ref{sec:training-effects}
examines longer training and separate tuning on smaller label heads.

\begin{table}[!ht]
  \centering\small
  \caption{RC-DS reconstruction differences from two baselines, reported as mean
  $\pm$ sample standard deviation across three paired seed summaries. For each checkpoint
  and seed, we first average its six task--sparsity differences. We then summarize the
  three seed averages, rather than treating 18 fits as independent observations.
  Positive values favor RC-DS. ``Hidden'' is the ordinary hidden-state SAE.
  The spread is conditional on the fixed checkpoint, texts, and training budget, not a
  confidence interval over models.
  Qwen2.5 supplied development data. Qwen3 and SmolLM2 were reserved for the semantic study.}
  \label{tab:reconstruction-summary}
  \begin{tabular}{lrrrr}
\toprule
Model & \shortstack{Gain-free EV\\vs. logits} & \shortstack{Gain-free EV\\vs. hidden} & \shortstack{Logit EV\\vs. logits} & \shortstack{Logit EV\\vs. hidden} \\
\midrule
Qwen2.5 & $+0.0216\pm0.0008$ & $+0.0043\pm0.0019$ & $-0.0058\pm0.0006$ & $+0.0052\pm0.0019$ \\
Qwen3 & $+0.0090\pm0.0012$ & $+0.0073\pm0.0011$ & $-0.0033\pm0.0010$ & $+0.0068\pm0.0008$ \\
SmolLM2 & $+0.0566\pm0.0025$ & $+0.0022\pm0.0006$ & $-0.0272\pm0.0011$ & $+0.0090\pm0.0017$ \\
\bottomrule
\end{tabular}

\end{table}

\subsection{Hidden-state and whitening controls explain the comparison}

Ordinary hidden-state SAEs are close competitors. RC-DS has higher mean reconstruction in
all 18 settings under both metrics, but its gain-free advantage over this baseline ranges
only from 0.0002 to 0.0124. On SmolLM2 Emotion at $K=8$, the hidden-state SAE reaches 0.6392,
compared with RC-DS at 0.6397. All three readouts retain the full hidden dimension.
Compact RC-DS and hidden coordinates therefore differ only by an orthogonal transformation.
The theory gives equivalent representations and transportable SAE solutions. These fits
were optimized separately with Adam in the two coordinate systems. Positive setting means
alone do not establish a reliable advantage. The paired seed ranges in
Table~\ref{tab:paired-hidden} show how these small differences vary across runs. In six
of the 18 gain-free comparisons, at least one paired seed favors hidden-state training
despite the positive mean.

Readout normalization also differs from generic variance normalization. In this full-vocabulary
grid, RC-DS exceeds standardization and ZCA on gain-free EV in all 18 setting means. It
exceeds ZCA on logit EV in all 18, but standardization in only 14. For SmolLM2 Emotion at
$K=8$, gain-free EV is 0.6397 for RC-DS, 0.4771 for standardization, and 0.2159 for ZCA.
ZCA is stronger in the longer-trained label-head comparison below and in many controlled
support-recovery settings. Readout normalization and covariance normalization favor different
objectives and training settings.

\subsection{Longer training changes the coordinate advantage}
\label{sec:training-effects}

Small constrained label heads expose the effect of training budget. The six-label Qwen2.5
and Qwen3 heads have condition numbers 1.425 and 1.353, respectively. Extending the same
Qwen2.5 SAEs from 60 to 1,000 epochs reduces the common-logit EV advantage from 0.0198 to
0.0016. None of the action-space fits meets the declared loss-stability rule at that cap.
ZCA reaches 0.9690 EV, above RC-DS at 0.9655 and centered logits at 0.9639.

A separate comparison uses independent encoder updates, new SAE seeds, and 360 new Emotion
prompts. Each method and sparsity level receives its own learning-rate selection on Qwen2.5,
with the selected rates transferred unchanged to Qwen3. At $K=4$, the RC-DS minus logit
differences are $+0.00058$ with seed interval $[-0.00011,+0.00128]$ on Qwen2.5 and $+0.00008$
with interval $[-0.00106,+0.00141]$ on Qwen3. At $K=1$ and $2$, centered-logit SAEs have
higher mean logit EV. Thirty-one of 180 fits still exceed the late-training loss-change
threshold. At $K=4$, both mean differences are near zero and both intervals span zero.
Appendix~\ref{app:prior-reconstruction} retains all coordinates, sparsity levels, and training
diagnostics. These small-label-head comparisons are separate from the full-vocabulary grid.

The comparisons identify three sources of variation in reconstruction: the readout and
data geometry, the target used to score errors, and the training procedure. They explain
the observed RC-DS gains without treating the coordinate choice as an independent source
of decision information.

\subsection{Comparing training at the same reconstruction targets}
\label{sec:common-quality}

We analyze the saved small-label-head fits using common fitting EV targets of 0.90, 0.95,
0.98, and 0.99, separately in logit and gain-free space. The analysis retains all 720 saved
checkpoints from the 180 confirmation fits. Learning rates remain fixed. For each target,
we select the first saved checkpoint that reaches it using fitting reconstruction only,
then score the selected model on the existing 360 held-out prompts. This is a post-hoc
analysis, not a new confirmation. Appendix~\ref{app:common-quality} gives the curves,
attained errors, probe procedure, and search costs.

At the original primary sparsity $K=4$, both RC-DS and centered logits reach every target
in all five seeds of both models. For gain-free targets, their first qualifying saved
checkpoints are identical in every pair. For logit targets, the checkpoints also match
except for two Qwen2.5 seed--target pairs, where logits qualify earlier. The saved resolution
therefore shows no earlier RC-DS checkpoint. Measurements exist only at 100, 500, 2,000,
and 6,000 updates, so they do not resolve speed differences between those checkpoints.

A shared threshold does not give identical fitting error. For example, at logit EV 0.90
on Qwen2.5, the selected RC-DS checkpoints have 0.0127 higher mean fitting EV, partly because
one RC-DS fit is selected later. Its held-out EV advantage of 0.0135 is consequently not a
matched-loss advantage. At the 0.99 logit target, held-out EV differences are 0.0012 on
Qwen2.5 and 0.0003 on Qwen3, with attained fitting-EV differences of about 0.0004 for each.
Single-feature label-detection differences at those checkpoints are $+0.0036$ and $-0.0038$
AUROC, respectively. These comparisons distinguish checkpoint cost, achieved reconstruction,
and feature usefulness without claiming exact loss matching or asymptotic equality.

\section{Applications: Label Detection and Steering}
\label{sec:feature-usefulness}

The preceding sections establish a training-geometry effect. We now ask what the learned
features add in two uses: detecting external labels and changing model behavior. These tests
evaluate the resulting features separately from reconstruction.

\subsection{Raw coordinates outperform RC-DS label detectors}
\label{sec:real-usefulness}

Using the full-vocabulary SAEs from Section~\ref{sec:reconstruction}, we select one feature
and its sign per class on 600 separate texts, then score AUROC on 1,200 texts and average
equally over classes. Labels are used for feature selection, not SAE training. Raw singular
coordinates receive the same selection procedure. They outperform the mean trained RC-DS
score in all 18 settings, with differences from 0.0002 to 0.1838 AUROC. The nine raw
model--task scores are each compared with two SAE sparsity levels, using three SAE seeds
per setting. Table~\ref{tab:raw-vs-trained} retains every comparison.
This is a descriptive result across nine model--task pairs and two SAE sparsities, not
eighteen independent model replications. The candidate pools differ: raw probes search the
retained rank, whereas each SAE supplies 1,024 features. In this test, sparse training does
not improve on selecting an existing coordinate, reinforcing the need for simple probe
baselines \citep{kantamneni2025useful}.

Multiplying a raw coordinate by a positive singular gain leaves its example rankings and
AUROC unchanged. Per-coordinate standardization also preserves rankings because it subtracts
a fixed mean and divides by a positive scale. The raw RC-DS, compact-logit, and standardized
probes therefore match exactly. Sparse
training learns new combinations of these coordinates, but its reconstruction advantage
does not translate into a stronger detector than the raw baseline in this test. Among
trained methods on the two reserved models, two setting means favor RC-DS by more than
0.01 AUROC, two favor logits, and eight have smaller differences. The fitting-data predictor
is less accurate than predicting zero difference. Appendix~\ref{app:full-readout-usefulness}
gives all scores, forecasts, and controls.

\subsection{Controlled steering and generation}
\label{sec:causal-boundary}

For constrained targets, the linear readout gives a state-specific oracle that maximizes
target probability under a per-example KL budget. An RC-DS intervention $\delta r$ is lifted
to the hidden state as $\delta h=VU^\top\delta r$. We compare learned directions with an
optimized fixed direction and random dictionaries given the same selection budget.

The frozen study using Qwen2.5-0.5B \citep{qwen2024qwen25} on the Emotion dataset
\citep{saravia2018carer} supplies a positive controlled case. Across five SAE seeds
on 360 balanced test examples, learned directions exceed their strongest control by
$+7.73\%$ of oracle gain, with seed-bootstrap interval $[+4.37\%,+11.18\%]$. All five seeds
are positive and the full confirmation rule passes. This compares learned directions with
direct and random controls. A separate longer-trained RC-DS--logit comparison on reused
development texts gives $-4.08\%$ of oracle gain, with interval $[-10.73\%,+0.82\%]$.

The other application tests give different outcomes. Bargaining's selected family closely
approximates direct steering, and Goofspiel's family has a smaller effect than its strongest
random control. A naturally active contextual bargaining variant misses its development
threshold. Two frozen Qwen sentiment-generation studies, fitted on IMDb or generated states,
also miss their confirmation rules, with both learned-minus-random intervals spanning zero.
Appendix~\ref{app:causal-boundary} retains the effect sizes, protocols, activity checks, and
uncertainty. These outcomes distinguish the established reconstruction effects from
feature usefulness for a particular target and state distribution.

\section{Discussion}
\label{sec:discussion}

The central finding is that sparse learning depends on how decision information is
represented, not only on which information is available. On the retained subspace, RC-DS
and centered logits determine each other. Ordinary squared-error training nevertheless
places different weights on their reconstruction errors. The exact wrapper makes this
distinction concrete: transforming the inputs, outputs, and loss together reproduces
RC-DS training from scaled coordinates. Reweighting the loss alone recovers only part of
the improvement. The resulting difference concerns both error weighting and training
parameterization, rather than additional decision information.

\paragraph{When removing readout scaling helps.}
In the base synthetic generator, sparse factors are formed before the readout applies its
gains. RC-DS restores that generating space, and its recovery advantage grows as the gains
become more unequal. Changing the generating space reverses the preferred coordinates at
the same condition number. A singular spectrum describes how strongly the readout scales
each direction, but does not specify how the data or its sparse factors occupy those
directions. This explains why condition number alone is insufficient. The relevant question
is whether removing the gains brings the training geometry closer to the structure being
recovered.

RC-DS provides a readout-based reference for answering that question. It removes gains
defined by the model's output map. Whitening instead uses the observed covariance, which
also reflects the data distribution. ZCA's higher support F1 in 34 of the 36 controlled
prediction pairs shows that removing readout gains is not the best preprocessing choice
in that comparison. Our predictor forecasts RC-DS relative to logits, not the best method
among all coordinate choices. The whitening results establish this distinction without
identifying which covariance property causes the advantage.

\paragraph{Reconstruction gains reflect both the target and the training procedure.}
Gain-free error treats retained directions equally, whereas logit error weights them by
the readout gains. The full-vocabulary trade-off is therefore meaningful: a method can
preserve more of the unscaled representation while preserving the output logits less
accurately. When the readout has full column rank, compact RC-DS coordinates are also a
rotation of the hidden state. Their small differences from independently trained
hidden-state SAEs concern optimization, not an increase in information or representational
capacity. Adam's sensitivity to the coordinate basis is relevant here
\citep{zhang2025adamrotations}, but our comparisons do not isolate it from every other
training choice.

Training budget answers a separate question. An advantage at equal updates describes
performance at that budget, not the best attainable reconstruction. Longer training and
separate tuning substantially reduce the gaps on the small label heads. The common-quality
analysis in Section~\ref{sec:common-quality} further separates the cost of reaching a
reconstruction target from performance at a fixed step count. These findings motivate
reporting learning curves and achieved errors alongside endpoint scores. They do not
establish the converged ordering for the separate full-vocabulary study.

\paragraph{Better reconstruction is not the same as better features.}
The reconstruction objective rewards representing the input with a sparse code. It does
not directly reward alignment with a particular semantic label or steering target. Raw
singular coordinates outperform mean trained RC-DS label detectors in all 18 tested
full-vocabulary settings. The gain-free reconstruction advantage over centered-logit SAEs
therefore does not make the trained features stronger label detectors than the raw baseline.
The controlled Qwen steering result establishes a useful learned basis for its tested
target and distribution. The game and generation controls show why that result does not
extend automatically to other uses. These are distinct evaluations of the same features,
not conflicting answers to one question.

\paragraph{Implications for SAE evaluation.}
Readout scaling should be an explicit experimental choice. Compare logits, gain-free
coordinates, hidden states, and covariance-based preprocessing under matched capacity and
tuning budgets. Score the same reconstructions in both gain-free and logit space, then
evaluate the intended use directly. Known-factor recovery needs ground-truth factors.
Label detection needs raw-coordinate controls, and steering needs direct and random
controls with matched selection budgets. This separates improvements in reconstruction
from evidence that sparse training has learned a more useful feature basis.

\section{Limitations}
\label{sec:limitations}

The case studies cover two games with three evaluation policies each and three fixed
language-model checkpoints, two reserved for semantic prediction. They are not a random
model survey. The 18-setting grid repeats tasks and sparsity levels within those checkpoints.
Seed summaries describe variation on fixed models and texts, not uncertainty across a
population of models. Reconstruction is a secondary analysis emphasized after inspecting
the results.

Several fits fail the declared loss-stability checks. The full-vocabulary comparison uses
a fixed 6,000-update budget, and the smaller-head training extensions do not determine its
converged outcome. Saved checkpoint spacing and unequal attained errors also prevent
interpreting the common-quality analysis as exact loss matching. The coordinate effects
have not been established across other SAE architectures.

Known-factor recovery is measured only in synthetic data. Real-model application tests
concern single-feature labels and the specified steering targets and state distributions.
The equivalence result holds on the retained subspace at a fixed checkpoint and mask.
Discarded directions and cross-checkpoint feature stability are outside that result.

\begin{samepage}
\section{Conclusion}
\label{sec:conclusion}

Readout scaling shapes sparse feature recovery without changing retained decision information.
The controlled studies show that the generating geometry changes which coordinates work best.
In the base generator, RC-DS improves recovery by removing readout distortion. Other
generating constructions favor retaining the gains. The condition-number sweep quantifies the
former case, and the exact wrapper isolates its training-geometry mechanism.
Real-model comparisons identify reconstruction
gains and explain their dependence on the target and training budget. Together, the method
and evidence make readout geometry an explicit, testable part of SAE design and evaluation.
Better reconstruction alone does not establish better recovered factors or more useful
features. Whitening and raw-coordinate controls are necessary to test those claims.

\label{main-text-end}
\par
\end{samepage}
\FloatBarrier
\section*{Ethics Statement}

The study uses synthetic data, simulated games, and public text corpora. It contains no newly
collected human-subject data. Sentiment continuations are machine-generated. The intervention
results concern small research models, not deployment-ready behavioral control. We retain failed
comparisons and distinguish development from confirmation to avoid overstating capability.

\begingroup
\raggedright
\bibliographystyle{plainnat}
\bibliography{references}
\endgroup

\clearpage
\appendix
\section*{Study overview}

Table~\ref{tab:study-overview} maps the reported evidence to its data, training budget, and
supporting appendix. The studies use different runs and budgets. In particular, the
20-seed condition-number sweep and the 12-seed coordinate-matching and loss studies are
separate, and the small-label-head training extensions do not extend the full-vocabulary fits.

Budgets are per SAE fit unless marked as policy training. An epoch is one pass through the
fitting data, whereas an update is one minibatch step. Language-model checkpoints remain
frozen. Seed counts refer to SAE or synthetic runs unless marked as policy seeds, and do
not count as independent language models. Data shared across studies and analyses that
reuse fitted models are identified below.

\begingroup
\footnotesize
\setlength{\tabcolsep}{3pt}
\setlength{\LTleft}{0pt}
\setlength{\LTright}{0pt plus 1fill}
\renewcommand{\arraystretch}{1.12}
\begin{longtable}{@{}
  >{\raggedright\arraybackslash}p{0.15\linewidth}
  >{\raggedright\arraybackslash}p{0.23\linewidth}
  >{\raggedright\arraybackslash}p{0.24\linewidth}
  >{\raggedright\arraybackslash}p{0.25\linewidth}
  >{\raggedright\arraybackslash}p{0.07\linewidth}@{}}
  \caption{Study overview. Data sizes describe fitting and evaluation roles, not independent
  model replications. The final column links to supporting appendices.}
  \label{tab:study-overview}\\
  \toprule
  Study & Models and data & Budget and seeds & Question tested & Details \\
  \midrule
  \endfirsthead
  \multicolumn{5}{@{}l}{\textbf{Table \thetable\ continued}}\\[3pt]
  \toprule
  Study & Models and data & Budget and seeds & Question tested & Details \\
  \midrule
  \endhead
  \midrule
  \multicolumn{5}{r@{}}{Continued on the next page}\\
  \endfoot
  \bottomrule
  \endlastfoot

  Gain and noise sweep &
  Synthetic, 16 factors in 8 dimensions, 4,096 samples &
  60 epochs, 20 seeds (200--219) &
  How does unequal scaling affect recovery when factors are generated before scaling? &
  \ref{app:protocols}, \ref{app:synthetic} \\
  \addlinespace[6pt]
  Approximate inverse loss &
  Separate synthetic samples, same dimensions and sample count as the sweep &
  60 epochs, 20 different seeds (300--319) &
  How much of the recovery difference does changing only the loss remove? &
  \ref{app:protocols}, \ref{app:synthetic} \\
  \addlinespace[6pt]
  Coordinate matching and attribution &
  Synthetic, 2,048 samples per run &
  40 epochs, 12 seeds per study, separate seed sets for four controls &
  Separate generating coordinates, exact wrapping, alternative training, and noise effects. &
  \ref{app:protocols}, \ref{app:synthetic} \\
  \addlinespace[6pt]
  Exact-wrapper check &
  Synthetic, condition numbers 3, 30, and 300 &
  Short optimizer trajectories, 3 seeds per condition (9 cases) &
  Do wrapped activations, gradients, parameters, and reconstructions match numerically? &
  \ref{app:formal} \\
  \addlinespace[6pt]
  Bargaining &
  OpenSpiel, 800 fitting and 400 evaluation episodes per policy &
  Policy: 3,000 episodes. SAE: 15 epochs. 3 development and 3 reused evaluation policy seeds &
  Compare logit reconstruction and decoder-family steering with direct and random controls. &
  \ref{app:protocols}, \ref{app:policy-reconstruction}, \ref{app:causal-boundary} \\
  \addlinespace[6pt]
  Goofspiel &
  OpenSpiel, 7 cards, 400 fitting and 200 evaluation episodes per policy &
  Policy: 5,000 episodes. SAE: 125 epochs. 3 development and 3 evaluation policy seeds &
  Test reconstruction and decoder-family steering on a second game. &
  \ref{app:protocols}, \ref{app:policy-reconstruction}, \ref{app:causal-boundary} \\
  \addlinespace[6pt]
  Contextual bargaining &
  3 development policies, 120 evaluation episodes each &
  SAE variants: 15 or 30 epochs. No reserved-policy evaluation &
  Can naturally active features capture the advantage of state-dependent control? &
  \ref{app:causal-boundary} \\
  \addlinespace[6pt]
  Coordinate comparison and extension &
  Qwen2.5 Emotion label head, 3,000 fitting and 480 reused development texts &
  60 epochs, then the same fits continued to 1,000. 5 paired seeds, 20 fits &
  Separate coordinate effects from training budget with every dictionary atom learned. &
  \ref{app:coordinate-training} \\
  \addlinespace[6pt]
  Encoder-update audit &
  Same Qwen2.5 fitting states as the coordinate comparison &
  2,000 epochs, 3 new seeds, 4 trainer variants, 48 fits &
  Check encoder copying, independent updates, and learning-rate schedules. &
  \ref{app:coordinate-training} \\
  \addlinespace[6pt]
  Separate tuning and fresh prompts &
  Qwen2.5 and Qwen3 label heads, 3,000 fitting and 360 new scoring texts &
  6,000 updates, 2 tuning seeds on Qwen2.5, 5 confirmation seeds per model &
  Compare reconstruction, sparsity, and compute after per-method tuning. Qwen3 reuses the selected rates. &
  \ref{app:fresh-tradeoffs} \\
  \addlinespace[6pt]
  Common fitting-quality comparison &
  Same two models, fits, and 360 scoring texts as the fresh-prompt study &
  No new training. 720 saved checkpoints from 180 fits, 5 seeds per setting &
  Compare first saved checkpoints reaching common fitting-EV targets, rather than only equal updates. &
  \ref{app:common-quality} \\
  \addlinespace[6pt]
  Correlation and sparsity controls &
  Synthetic, 2,048 fitting and 512 scoring samples &
  3,000 updates, 3 seeds, 2 rates, 288 fits &
  How do factor correlation and simultaneous activity change directional recovery? &
  \ref{app:fresh-tradeoffs} \\
  \addlinespace[6pt]
  Frozen recovery predictor &
  12 new synthetic settings, 2,048 fitting and 512 scoring samples each &
  6,000 updates, 3 fresh seeds per setting, 288 fits &
  Can fitting-data geometry predict changes in active-factor support F1? &
  \ref{app:observable-recovery} \\
  \addlinespace[6pt]
  Constrained semantic probes &
  Qwen2.5 and Qwen3 label heads, 3,000 fitting and 480 new scoring texts &
  6,000 updates, 3 seeds, 2 rates, 96 fits &
  Does the coordinate choice improve single-feature Emotion label detection? &
  \ref{app:observable-recovery} \\
  \addlinespace[6pt]
  Qwen full-readout comparison &
  Qwen2.5 vocabulary head, 3,000 fitting, 240 tuning, 240 scoring Emotion states, reused data &
  400 epochs, 2 seeds, 2 rates, 8 methods, 32 fits &
  Compare both reconstruction metrics with hidden-state, whitening, and loss-only controls. &
  \ref{app:qwen-full-readout} \\
  \addlinespace[6pt]
  Full-readout, three-model study &
  Qwen2.5, Qwen3, SmolLM2. Emotion, AG News, IMDb. Per task: 2,400 fit, 600 select, 1,200 score &
  6,000 updates, 3 seeds, 2 rates, 2 sparsities, 540 fits &
  Test semantic forecasts on reserved models. The same fits supply the 18-setting reconstruction analysis. &
  \ref{app:full-readout-usefulness} \\
  \addlinespace[6pt]
  Constrained Emotion steering &
  Qwen2.5 label head, 3,000 fitting, 480 development, 360 test texts &
  60 epochs, 5 seeds, anchored SAEs &
  Do active learned features beat fixed and random controls under a shared KL budget? &
  \ref{app:causal-boundary} \\
  \addlinespace[6pt]
  IMDb-to-generation steering &
  Qwen2.5, 3,072 fitting reviews, 1,024 association reviews, 15 test prefixes &
  60 epochs, 5 seeds. Five 32-token samples per prefix and polarity &
  Does sentiment steering learned from review states transfer to free generation? &
  \ref{app:causal-boundary} \\
  \addlinespace[6pt]
  Generated-state steering &
  Qwen2.5, OpenWebText prompts: 128 fit, 64 associate, 64 test &
  60 epochs, 5 new seeds. Three 32-token samples per test prompt and polarity &
  Does fitting and selecting features on generated states improve sentiment steering? &
  \ref{app:causal-boundary} \\
  \addlinespace[6pt]
  Sequence-aware follow-up &
  Qwen2.5, 64 further OpenWebText prompts &
  Reuses one generated-state SAE seed. Three 16-token samples per prompt and polarity &
  Does sequence-aware feature selection improve steering on new development prompts? &
  \ref{app:causal-boundary} \\
\end{longtable}
\endgroup
\clearpage

\section{Mathematical Details}
\label{app:formal}

\subsection{Centering allowed actions and choosing the rank}

For a legal set $L$, the selector $S_L$ removes the rows of the global action head that cannot be
executed and the centering matrix $C_L=I-|L|^{-1}\ones\ones^\top$ subtracts the mean legal
logit. Because $C_L\ones=0$, adding the same scalar to every legal logit leaves both the
centered logits and their softmax probabilities unchanged. The centered legal head
$W_c=C_LS_LW$ therefore contains every linear direction that can change the immediate policy over
legal actions. Hidden directions in its nullspace cannot change those probabilities.

Let $W_c^{(\rho)}=U\Sigma V^\top$ retain singular values above the declared numerical tolerance.
Writing $P_U=UU^\top$ and $P_V=VV^\top$, the polar representation is
$r=UV^\top(h-\mu)$. Its image lies in $\operatorname{col}(U)$ and its squared norm is
\begin{equation}
  \lVert r\rVert_2^2
  =(h-\mu)^\top P_V(h-\mu),
\end{equation}
so it preserves Euclidean length within the retained row space while deleting the orthogonal
nullspace. The forward and inverse maps in Remark~\ref{rem:readout-equivalence} act only on
this retained subspace.
If $E=W_c-W_c^{(\rho)}$, the difference between the original and retained centered-logit response
is bounded by
\begin{equation}
  \lVert Eh\rVert_2\leq \lVert E\rVert_2\lVert h\rVert_2.
\end{equation}
We record the tolerance and rank for each allowed-action set rather than assuming
$\rho=|L|-1$.

\subsection{Proof details for fixed-mask equivalence}

For fixed checkpoint and mask, suppress the indices and define
$r=UV^\top(h-\mu)$. Since $U^\top U=V^\top V=I_\rho$,
\begin{align}
  U\Sigma U^\top r
  &=U\Sigma U^\top UV^\top(h-\mu)\\
  &=U\Sigma V^\top(h-\mu)\\
  &=W_c^{(\rho)}(h-\mu).
\end{align}
Adding $W_c^{(\rho)}\mu+c$ yields the rank-truncated centered logits. Conversely, the centered
residual belongs to $\operatorname{col}(U)$, and therefore
\begin{align}
  U\Sigma^{-1}U^\top
  \left(W_c^{(\rho)}(h-\mu)\right)
  &=U\Sigma^{-1}U^\top U\Sigma V^\top(h-\mu)\\
  &=UV^\top(h-\mu)=r.
\end{align}
The map is affine because the fitting mean and actor bias appear as fixed offsets. It is invertible
only between the retained representations. Neither side recovers hidden-state components in
$\ker(W_c^{(\rho)})$.

\subsection{Hidden-basis invariance and its limit}

For an orthogonal hidden-state reparameterization $h'=Oh$ and $W'=WO^\top$, a valid right
singular basis is $V'=OV$. Hence
\begin{equation}
  U{V'}^\top(h'-O\mu)=UV^\top O^\top O(h-\mu)=UV^\top(h-\mu).
\end{equation}
The same cancellation handles simultaneous sign changes and orthogonal rotations inside a
degenerate singular subspace. This statement does not extend to an arbitrary nonsingular
reparameterization $T$: Euclidean polar factors generally change under $h'=Th$ and
$W'=WT^{-1}$. Centered logits have the stronger general-linear invariance. We therefore interpret
RC-DS as one choice of how to measure distances, not a representation independent of all
choices of hidden-state coordinates.

\subsection{Why the exact wrapper reproduces RC-DS training}

For one fixed allowed-action set, write the scaled coordinates as
\begin{equation}
  y_\alpha=T_\alpha r,
  \qquad
  T_\alpha=U\Sigma^\alpha U^\top
  \quad\text{on }\operatorname{col}(U).
\end{equation}
Let an SAE in RC-DS coordinates be $F_\theta(r)$. The exact wrapper for scaled inputs is
\begin{equation}
  \widehat y_\alpha
  =T_\alpha F_\theta(T_\alpha^{-1}y_\alpha),
  \qquad
  \mathcal L_\alpha
  =\left\lVert T_\alpha^{-1}
  (y_\alpha-\widehat y_\alpha)\right\rVert_2^2.
\end{equation}
Substitution gives
$\mathcal L_\alpha=\lVert r-F_\theta(r)\rVert_2^2$. Thus the forward values, sparse activations,
gradients, and optimizer trajectory are identical when the wrapper transforms the encoder input,
decoder output, and loss consistently. The numerical test spans condition numbers 3, 30, and 300
and observes a maximum discrepancy of $1.33\times10^{-15}$.

In the policy comparison that changes only the loss, the encoder still receives $y_\alpha$.
Only the error weighting changes. The inverse transform acts as the identity on
$\operatorname{col}(U)^\perp$ so that arbitrary decoder residuals receive finite loss:
\begin{equation}
  \widetilde T_\alpha^{-1}
  =U\Sigma^{-\alpha}U^\top+(I-UU^\top).
\end{equation}
With Top-$K$ selection and state-dependent masks, this is not the same model as the complete
wrapper. Changing only the loss therefore does not guarantee the same training trajectory as RC-DS.

\section{Experimental Protocols}
\label{app:protocols}

\subsection{Common design and evaluation}

We evaluate generalization across policies, not by counting their many transitions as independent
policies. Complete episodes are assigned to separate SAE-training, method-development, and
evaluation sets. The fitting mean,
readout factors, RMS scales, and SAE parameters use only the fitting partition. Development
outcomes may select a trainer or intervention family. Evaluation on a previously inspected policy
is labeled post-hoc validation rather than confirmation.

The policy comparisons use $\alpha\in\{0,\tfrac12,1\}$ with the same SAE capacity. Reconstruction
is scored as centered-logit explained variance in a shared output space, using
Equation~\ref{eq:explained-variance}.
The synthetic recovery score first maps learned decoder directions back to the generating
coordinates. For learned directions $\hat d_i$ and known factors $d_j$, it is
\begin{equation}
  R=\max_{\pi}\frac{1}{m}\sum_{i=1}^{m}
  \frac{\hat d_i^\top d_{\pi(i)}}{\|\hat d_i\|_2\|d_{\pi(i)}\|_2},
\end{equation}
where $\pi$ is a one-to-one assignment and $m=16$ in the recovery studies. As in the
implementation, cosine normalization uses an epsilon floor for a zero-norm direction.
Cosines are not made absolute: the generator and SAE use nonnegative codes, so reversing
a direction is not treated as recovering the same factor. The score measures directional
agreement, not the fraction of factors recovered exactly or a reconstruction score.

Causal rollout conditions reuse environment, policy-action, and opponent-action random streams.
We report the target statistic, conditional non-target KL, return, and policy agreement. Random
controls receive the same dictionary width, family size, and outcome-blind selection procedure as
the SAE. Direct controls apply the corresponding centered action-space direction without an SAE.

\begin{table}[t]
  \centering
  \small
  \caption{Policy settings fixed before evaluation. Each policy is tested with no readout
  scaling, square-root scaling, and full scaling.}
  \label{tab:app-policy-protocols}
  \begin{tabular}{lcc}
    \toprule
    Setting & Bargaining & Goofspiel \\
    \midrule
    Development policy seeds & 710--712 & 920--922 \\
    Evaluation policy seeds & 810--812 & 1120--1122 \\
    Policy-training episodes & 3,000 & 5,000 \\
    Hidden layers & $128,128$ & $128,128$ \\
    SAE fit/evaluation episodes & $800/400$ & $400/200$ \\
    Dictionary size / Top-$K$ & $192/8$ & $192/8$ \\
    SAE epochs / batch size & $15/512$ & $125/512$ \\
    SAE learning rate & $10^{-3}$ & $10^{-3}$ \\
    Selected family size & 24 & 2 \\
    Random dictionaries & 8 & 8 \\
    Rollout episodes per cell & 500 & 500 \\
    \bottomrule
  \end{tabular}
\end{table}

\subsection{Sparse model and selection of steering directions}

The policy SAEs use a nonnegative Top-$K$ encoder with a 192-atom decoder and eight active atoms
per observation. This implementation includes one fixed direction per action, called an
\emph{anchor}, and learns the remaining atoms. Bargaining penalizes similarity between decoder
atoms with a coherence weight of 0.05. Groups of steering directions are selected
by orthogonal least squares using only decoder directions, the fixed readout, and the frequencies
of allowed-action sets in the SAE training data. Rewards and rollout outcomes
are not inputs to family selection.

The resulting intervention is a signed linear combination of selected decoder atoms. It is a test
of whether a few dictionary directions can produce a useful intervention. It does not establish
that one naturally active feature represents one meaningful concept.

\subsection{Synthetic generator and uncertainty}

The generator samples 16 normalized Gaussian directions in eight dimensions, chooses two active
factors per observation, and assigns nonnegative coefficients uniformly between 0.5 and 1.5.
The signal is their weighted sum. The gain sweep applies diagonal log-spaced scales between
1 and $1/\kappa$, raised to $\alpha$, and centers each observed dataset before training.
The initial implementation check uses five seeds and 4,096 samples at condition number 100.
The invertibility study uses seeds 200--219, 4,096 samples, 60 epochs, and condition numbers
$1,3,10,30,100$, with no noise, pre-gain noise, and post-gain noise. Noise has standard
deviation 0.01 before its designated transformation. Pre-gain noise is transformed with the
signal, preserving information across the scaling choices. Post-gain noise is added after the
transformation instead.
Equivalent effective spectra share a training block, including the identical RC-DS
inputs across nominal condition numbers, and are not additional independent replications.

The approximate inverse-loss comparison uses a separate set of seeds, 300--319, with
4,096 samples and 60 epochs at condition numbers 10, 30, and 100. It tests no noise and
pre-gain noise. Its RC-DS and ordinary scaled-input baselines are trained on these new
sampled problems, not reused from the seeds 200--219 sweep.

The follow-up comparisons use 2,048 samples, 40 epochs, and twelve seeds per test: 400--411 for
the loss and wrapper comparison, 420--431 for the alternative trainer, 440--451 for changes to
the generating coordinates, and 460--471 for noise and scaling tests. The last test adds condition number
300. Both the sweep and campaign use batch size 512 and learning rate $10^{-3}$.
Reported intervals use paired bootstrap resampling over seeds: 10,000 resamples for the sweep
and 5,000 for the campaign. The nine-case numerical equivariance check is a separate study of
short training runs, not another set of factor-recovery measurements.

\section{Additional Synthetic Results}
\label{app:synthetic}

\begin{table}[!ht]
  \centering
  \small
  \caption{Factor recovery at condition number 100 under the invertibility and noise-placement
  controls, using seeds 200--219, 4,096 samples, and 60 epochs. Intervals are paired 95\%
  bootstrap intervals for $\alpha=0$ minus $\alpha=1$. Means and differences are computed
  at full precision and rounded separately.}
  \label{tab:app-noise}
  \begin{tabular}{lrrrr}
    \toprule
    Noise & $\alpha=0$ & $\alpha=1/2$ & $\alpha=1$ & Difference [95\% CI] \\
    \midrule
    None & 0.9134 & 0.7435 & 0.5063 & 0.4070 [0.3702, 0.4446] \\
    Before gain & 0.9135 & 0.7464 & 0.5101 & 0.4034 [0.3627, 0.4445] \\
    After gain & 0.9135 & 0.7342 & 0.4876 & 0.4259 [0.3958, 0.4556] \\
    \bottomrule
  \end{tabular}
\end{table}

Mean recovery is tied at condition number one and separates as gains become more unequal in
this generator and finite-budget sweep. This is not a guarantee of monotonic change in every
seed or dataset. The pre-gain row is the crucial control: because the same invertible map is
applied to signal and noise, its recovery gap cannot be attributed to additional information in
the RC-DS arm.

The separate approximate inverse-loss comparison uses seeds 300--319, rather than
200--219 in Table~\ref{tab:app-noise}. Its baseline means therefore come from different
sampled problems. The inverse-weighted loss recovers part of the difference but not all of it.
At condition number 100 without noise, recovery is 0.9131 for MSE in RC-DS coordinates, 0.6531
for the approximate inverse loss, and 0.5158 for ordinary MSE in scaled coordinates. With pre-gain noise, the
corresponding values are 0.9063, 0.6484, and 0.5085. The inverse-minus-ordinary differences are
0.1373 [0.1062, 0.1739] and 0.1399 [0.1052, 0.1792]. These approximate results motivated the exact
wrapper rather than serving as evidence against mathematical equivalence.

In the loss and wrapper comparison, RC-DS and the exact wrapper match in every trial, with
mean recovery 0.8176, compared with 0.6284 for the approximate inverse loss and 0.4841 for
ordinary MSE in scaled coordinates. The exact-wrapper advantage over this baseline is 0.3335
[0.2851, 0.3816]. A separate untuned trainer yields 0.6359 versus 0.4703, a difference of 0.1656
[0.1377, 0.2008]. When generating factors are moved among the three coordinate systems, recovery
is highest on the matched diagonal in every row (0.8364). Matching cancels the transform and
returns the same training inputs, initialization, and minibatch schedule. Equal diagonal
scores are therefore a construction check. The off-diagonal losses establish the
coordinate-matching effect within this generator (Section~\ref{sec:coordinate-matching}).

These controls support a specific empirical claim: readout gains can change recovery under
ordinary sparse training without changing the available information. They do not identify one
universally best coordinate system, demonstrate equal convergence across coordinates, or
predict that removing readout scaling must improve output-weighted reconstruction in real models.

\FloatBarrier
\section{Policy Reconstruction}
\label{app:policy-reconstruction}

Bargaining uses an imperfect-information negotiation task. Its three evaluation policies were
inspected during method development, so reconstruction and corrected causal rollouts are post-hoc
validation. Goofspiel uses seven cards, descending prize cards, and point-difference returns.
Its evaluation seeds are fresh under the retained protocol. Both environments train against a
uniform opponent. Episode partitions, model sizes, and budgets appear in
Table~\ref{tab:app-policy-protocols}.

\begin{table}[!ht]
  \centering
  \small
  \caption{Bargaining reconstruction by policy. All values are evaluation centered-logit EV.}
  \label{tab:app-bargaining-ev}
  \begin{tabular}{lrrr}
    \toprule
    Policy & $\alpha=0$ & $\alpha=1/2$ & $\alpha=1$ \\
    \midrule
    810 & 0.9108 & 0.9107 & 0.8739 \\
    811 & 0.9207 & 0.9167 & 0.9009 \\
    812 & 0.9116 & 0.8774 & 0.8657 \\
    \midrule
    Macro mean & 0.9144 & 0.9016 & 0.8802 \\
    \bottomrule
  \end{tabular}
\end{table}

Goofspiel evaluation means are 0.8637, 0.8181, and 0.7654 for $\alpha=0,1/2,1$.
The RC-DS minus centered-logit differences for seeds 1120--1122 are
$-0.0045,+0.1085,+0.1909$. Thus, the higher mean does not imply improvement on every policy.
The inverse-gain-loss-only method reaches 0.5607 in bargaining and 0.2384 in Goofspiel. It still
feeds scaled inputs to the encoder, with scaling that depends on the allowed actions, unlike the exact
wrapper. Means are not pooled across environments, and three policy seeds do not support a
population-level significance claim.

\FloatBarrier
\section{Coordinate Comparisons and Training Checks}
\label{app:coordinate-training}

\subsection{Qwen comparison with every dictionary atom learned}

The three training comparisons below use the same 3,000 fitting and 480 already-opened Qwen Emotion development
states. They do not reuse the earlier Qwen confirmation test data. The centered six-label head has
rank five and condition number 1.425. Hidden states, centered logits, RC-DS, and ZCA-whitened
logits receive 64 trainable atoms, Top-$K$ 4, no fixed anchors, and fitting-only means, scales,
and whitening statistics. The three action-space arms have 838 parameters each. The hidden-state
arm has 115,648. Thus, dictionary width is matched, but the hidden-state comparison is not
parameter matched. Seeds 262001--262005 share initialization rows and minibatch schedules.

The 60-epoch comparison copies decoder weights into the encoder after each update and retains
all twenty outcomes. Known labels guide the intervention for each example. This is a controlled
steering test, not label prediction or free generation. Controls are the maximum permitted
target-probability increase under a per-example KL limit (the oracle), an optimized fixed
direction, and eight shared random dictionaries. Random controls match each SAE's actual number
of active features. At 60 epochs, the paired RC-DS minus centered-logit effect is $+0.52\%$ of oracle gain,
with seed interval $[-0.14\%,+1.19\%]$ and prompt interval $[-0.58\%,+1.64\%]$. Neither
interval establishes a coordinate-specific steering benefit.

\subsection{Longer training without evaluation-based stopping}

The extension reproduces every 60-epoch parameter and fitting-history value exactly before
continuing the same optimizer and shuffle states. Beginning at epoch 200, checks occur every
100 epochs and compare successive 50-epoch mean fitting MSE and total objective. Both absolute
relative changes must be below 1\% for three successive checks, for all twenty fits, or training
stops at a common 1,000-epoch cap. The stopping decision is recorded before development scoring.
The cap is reached: only the five hidden-state fits satisfy the stability condition. All fifteen
action-space fits remain outside it, so these are not converged-model comparisons.

\begin{table}[!ht]
  \centering
  \small
  \caption{Qwen coordinate comparison on reused development data. EV is measured in common
  centered-logit space. Excess is learned minus strongest control gain, divided by oracle gain,
  after 1,000 epochs. All entries average the same five seeds.}
  \label{tab:app-coordinate-training}
  \begin{tabular}{lrrrr}
    \toprule
    Coordinates & EV, 60 epochs & EV, 1,000 & Excess, 1,000 & Stable fits \\
    \midrule
    Hidden states & 0.7735 & 0.8025 & $+8.42\%$ & 5/5 \\
    Centered logits & 0.6066 & 0.9639 & $+7.14\%$ & 0/5 \\
    RC-DS & 0.6264 & 0.9655 & $+0.15\%$ & 0/5 \\
    ZCA-whitened logits & 0.7041 & 0.9690 & $+5.07\%$ & 0/5 \\
    \bottomrule
  \end{tabular}
\end{table}

At the cap, the direct paired RC-DS minus centered-logit steering contrast is $-4.08\%$ of
oracle gain: its seed interval is $[-10.73\%,+0.82\%]$ and its prompt interval is
$[-5.42\%,-2.77\%]$. These estimate different sources of uncertainty. The negative prompt
interval for a fixed seed panel does not establish a seed-robust effect. The mean effective
nonzero counts are 3.995, 3.939, 3.955, and 3.970 in table order. Each arm's random control
uses its actual active support. The raw RC-DS minus centered-logit target-probability difference
is $-0.004910$. The shared oracle gain is measured from baseline probability 0.3494 to oracle
probability 0.4697. No failed seed is discarded.

\subsection{Checking encoder updates and learning-rate schedules}

\begin{figure}[!ht]
  \centering
  \includegraphics[width=\linewidth]{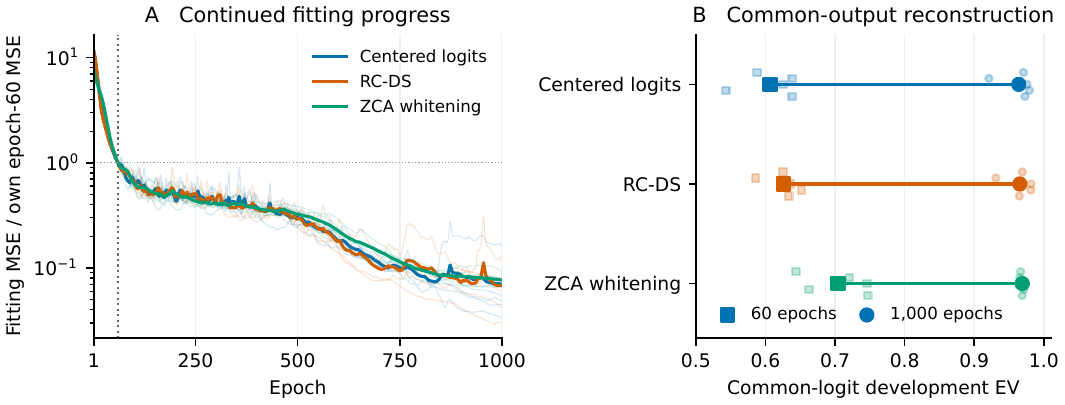}
  \caption{Training-budget sensitivity on the Qwen2.5 label head, using the same five paired
  seeds and three parameter-matched action-space arms. Left: fitting histories normalized
  within each fit to epoch 60. Thin curves show individual seeds and thick curves their means.
  Right: common-logit development EV at the two measured endpoints. Faint points show every
  seed and large markers show means. Connecting lines do not imply intermediate measurements.
  The extension reaches its cap, not convergence.}
  \label{fig:training-budget}
\end{figure}

The copy-based trainer computes an encoder gradient but overwrites the updated encoder weight
with the normalized decoder after each Adam step. A one-step probe has encoder-gradient norm
0.3193, yet setting the encoder-weight rate to zero changes no final parameter under copying.
Without copying, the maximum parameter difference is 0.001000. The copy/constant branch
reproduces the recorded trainer exactly over 60 epochs. Finite differences validate the
independent encoder away from Top-$K$ ties.

A comparison fixed before evaluation tests all four combinations of copied or independently
learned encoder weights and constant or cosine-decayed learning rates. Every representation uses seeds 262101--262103, 2,000
epochs, batch size 512, coherence weight 0.001, encoder/base rate 0.001, and three times that
rate for the decoder. Decay stops at 1\% of the initial rate. A training procedure is selected
only if all twelve fits meet the predeclared stability and method-specific accuracy requirements.
New-prompt and steering outcomes are not used for selection.

\begin{table}[!ht]
  \centering
  \small
  \caption{The 48-fit training audit. No recipe satisfies both conditions for all twelve fits,
  so no recipe is selected and the twenty reserved-seed confirmation fits are not run.}
  \label{tab:app-trainer-audit}
  \begin{tabular}{lrrr}
    \toprule
    Encoder / learning rate & Stable fits & Accurate fits & Both \\
    \midrule
    Copy / constant & 2/12 & 12/12 & 2/12 \\
    Copy / decay & 10/12 & 3/12 & 3/12 \\
    Independent / constant & 9/12 & 12/12 & 9/12 \\
    Independent / decay & 11/12 & 11/12 & 10/12 \\
    \bottomrule
  \end{tabular}
\end{table}

\begin{table}[!ht]
  \centering
  \small
  \caption{Effect of learning encoder weights independently at a constant learning rate.
  MSE is measured on training data in each method's normalized input coordinates.
  Compare MSE within rows, not between coordinate systems.
  EV uses the common logit space, also on fitting data.}
  \label{tab:app-encoder-update}
  \begin{tabular}{lrrrr}
    \toprule
    Coordinates & Copy MSE & Independent MSE & Copy EV & Independent EV \\
    \midrule
    Hidden states & 0.3774 & 0.3058 & 0.8002 & 0.8656 \\
    Centered logits & 0.01069 & 0.006251 & 0.9894 & 0.9937 \\
    RC-DS & 0.01425 & 0.006699 & 0.9865 & 0.9935 \\
    Whitened logits & 0.01463 & 0.01246 & 0.9852 & 0.9873 \\
    \bottomrule
  \end{tabular}
\end{table}

Independent encoder learning improves training fit in every method. A small learning rate can
make loss nearly constant even when the fit is poor, so stability alone does not establish accuracy.
The independent-encoder, decaying-rate procedure
misses in two fits: a centered-logit fit exceeds the 1\% change threshold while still improving,
and a whitened fit exceeds its MSE ceiling by 0.727\%. The thresholds are not relaxed. These
failures do not establish harmful oscillation in every case or rule out other training recipes.
The audit ends without selecting a trainer, extracting fresh model states, or scoring fresh
interventions. Earlier results remain attributable to their original algorithms and budgets.

Full per-epoch histories, trained parameters, optimizer state, coordinate maps, and checksums
remain in the internal research record. Execution retries and the unused fresh-panel scoring
correction are documented there. Neither constitutes additional independent evidence.

\FloatBarrier
\section{Separate Tuning and Evaluation on New Qwen Prompts}
\label{app:fresh-tradeoffs}

This comparison fixes its protocol before training the new SAEs or evaluating new prompts. It uses
the same 3,000 Qwen2.5 training states as the coordinate comparison, verified by checksum. Qwen3 states are
extracted for those same texts. The pinned model revisions are:

\begin{tabular}{@{}ll@{}}
  Qwen2.5 & \texttt{060db6499f32faf8b98477b0a26969ef7d8b9987} \\
  Qwen3 & \texttt{da87bfb608c14b7cf20ba1ce41287e8de496c0cd}
\end{tabular}

The extraction path is checked
against the complete model's final hidden states and label logits. Both models use the same
six distinct single-token labels and prompt template. The two checkpoints belong to the same
model family and perform the same task. They do not represent independent model families.

Each action-space SAE has six inputs and 838 parameters. PCA coordinates are padded to the
same width. The hidden references have 115,648 and 132,160 parameters, respectively, with the
same 64-atom dictionary. They also retain directions that cannot affect the readout. Neither
parameter count nor dictionary size relative to input size is matched to the action-space SAEs. All normalization statistics use fitting
rows only. Numerical rank uses relative tolerance $10^{-6}$, excluding the centering null
direction. Whitening uses a relative covariance floor of $10^{-6}$, which is inactive on the
five retained covariance eigenvalues of both label heads. These are full retained-rank PCA
and ZCA controls, not truncated dimensionality-reduction baselines.

Initial decoder directions use paired fitting-row indices, with one fitting RMS scale per arm.
The encoder initializes from the decoder and then learns independently. Adam updates all
parameters using MSE in each method's input coordinates, with gradient clipping at norm one and no coherence
penalty or post-update encoder copying. Minibatches sample with replacement. Learning rates
decay linearly to one tenth of their initial value by update 4,800 and remain fixed through
update 6,000. The common candidate rates are $0.0003,0.001,0.003$, with tuning seeds
270101--270102. Every method and $K$ chooses its own rate using final tuning logit EV.
Five confirmation seeds 270201--270205 then use those rates. Qwen3 receives no additional
learning-rate search. Earlier curve checkpoints use the final-selected rate too, so their
search cost must not be treated as zero.
Equal search budgets do not establish equally optimized models. Several choices are at the
largest tested learning rate, and better settings outside this grid are not ruled out.

The new evaluation set contains 360 label-balanced examples from the pinned Emotion training split at sampling
seed 270299. It excludes fitting texts and the entire official validation and test splits after
normalized-text deduplication. The pipeline also checks identity after each tokenizer's prompt
formatting and truncation. All trained SAEs are saved and checksummed before this set is
evaluated. Evaluation proceeds regardless of the sign of the result or training accuracy.
The earlier training audit's reserved set remains unused. These examples are new to this
evaluation protocol, but may have appeared in the language models' unknown pretraining data.

Saved checkpoints at 100, 500, 2,000, and 6,000 updates provide the learning curves. Every
100 updates, full-fitting MSE is recorded. The late-change diagnostic compares the means of
the final two five-checkpoint windows. A change below 1\% is descriptive, not proof of a global
optimum. Training time uses one CPU thread and excludes initialization, extraction,
preprocessing, evaluation, and serialization. Dense matrix-multiplication work is approximated
by $12Tbdm$ for $T$ updates, batch size $b$, input width $d$, and dictionary width $m$.
It excludes optimizer, normalization, and Top-$K$ overhead. Timings are not end-to-end speedups.

\begin{table}[!ht]
\centering\small
\caption{Centered-logit EV on 360 fresh prompts at 6,000 updates. Means of all five SAE seeds. Hidden is a larger, unmatched reference.}
\begin{tabular}{llrrrrrr}
\toprule
Model & $K$ & Logits & RC-DS & Std. & ZCA & PCA & Hidden \\
\midrule
Qwen2.5-0.5B & 1 & 0.9203 & 0.9175 & 0.9178 & 0.8560 & 0.8425 & 0.7010 \\
Qwen2.5-0.5B & 2 & 0.9437 & 0.9398 & 0.9386 & 0.9175 & 0.9210 & 0.7491 \\
Qwen2.5-0.5B & 4 & 0.9920 & 0.9926 & 0.9917 & 0.9866 & 0.9839 & 0.8120 \\
Qwen3-0.6B-Base & 1 & 0.9228 & 0.9190 & 0.9161 & 0.8540 & 0.8397 & 0.7705 \\
Qwen3-0.6B-Base & 2 & 0.9409 & 0.9344 & 0.9323 & 0.9142 & 0.9157 & 0.7987 \\
Qwen3-0.6B-Base & 4 & 0.9918 & 0.9919 & 0.9916 & 0.9838 & 0.9819 & 0.8440 \\
\bottomrule
\end{tabular}
\end{table}

\begin{table}[!ht]
\centering\small
\caption{Gain-free EV on 360 fresh prompts at 6,000 updates. Means of all five SAE seeds. Hidden is a larger, unmatched reference.}
\begin{tabular}{llrrrrrr}
\toprule
Model & $K$ & Logits & RC-DS & Std. & ZCA & PCA & Hidden \\
\midrule
Qwen2.5-0.5B & 1 & 0.9090 & 0.9126 & 0.9055 & 0.8537 & 0.8410 & 0.6951 \\
Qwen2.5-0.5B & 2 & 0.9372 & 0.9373 & 0.9313 & 0.9169 & 0.9201 & 0.7448 \\
Qwen2.5-0.5B & 4 & 0.9910 & 0.9923 & 0.9905 & 0.9865 & 0.9838 & 0.8084 \\
Qwen3-0.6B-Base & 1 & 0.9176 & 0.9182 & 0.9070 & 0.8539 & 0.8401 & 0.7691 \\
Qwen3-0.6B-Base & 2 & 0.9371 & 0.9335 & 0.9262 & 0.9144 & 0.9158 & 0.7985 \\
Qwen3-0.6B-Base & 4 & 0.9910 & 0.9916 & 0.9907 & 0.9837 & 0.9818 & 0.8446 \\
\bottomrule
\end{tabular}
\end{table}

\begin{figure}[!ht]
  \centering
  \includegraphics[width=\linewidth]{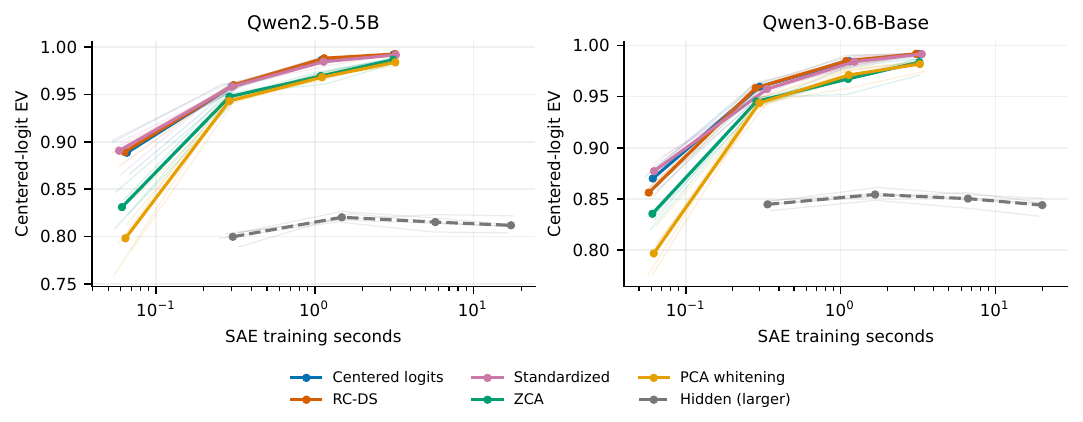}
  \caption{Logit EV on new prompts versus SAE training time at $K=4$. Points show the four
  checkpoints selected before evaluation. Thin lines show each seed and thick lines show means.
  The hidden-state SAE is larger. Tuning adds to the reported time, and Qwen3 reuses Qwen2.5's
  rates. Lines connect measurements without implying scores at intermediate times.}
\end{figure}

\begin{figure}[!ht]
  \centering
  \includegraphics[width=\linewidth]{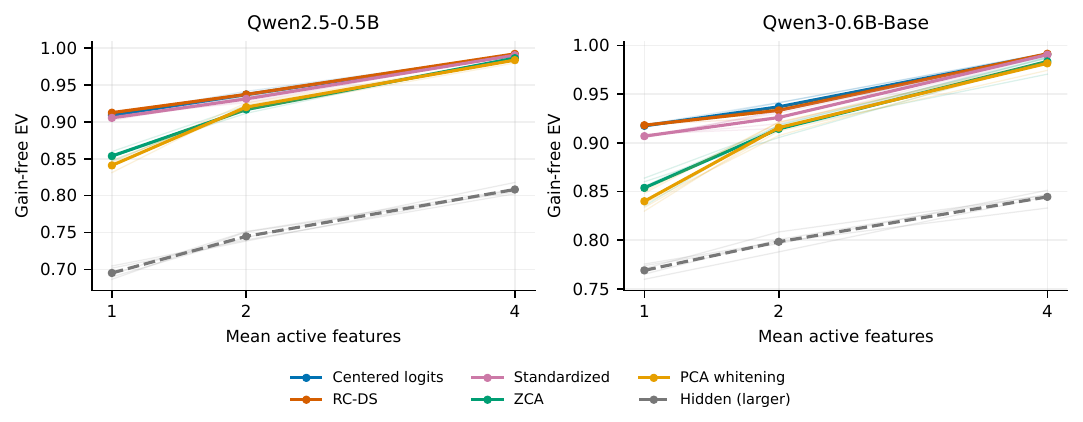}
  \caption{Gain-free reconstruction for the same fresh data, fitted models, and sparsity levels
  as the common-logit comparison. These metrics score the same residuals with different
  weighting. Neither identifies ground-truth sparse factors in the language models.}
\end{figure}

\FloatBarrier
\subsection{Common fitting-quality thresholds and checkpoint cost}
\label{app:common-quality}

This post-hoc analysis reuses the complete small-label-head comparison in this appendix.
It does not retrain SAEs, change learning rates, or open new prompts. Both models, all six
coordinate methods, $K\in\{1,2,4\}$, five seeds, and four saved checkpoints are retained.
All 720 checkpoints are loaded from their recorded files, and all four original held-out
reconstruction metrics are reproduced within absolute tolerance $10^{-10}$. Consumed source
files are checked against their recorded hashes and checked again after replay.

For each fit, we compute common fitting logit EV and gain-free EV using the original
fitting means and coordinate maps. Each metric has targets 0.90, 0.95, 0.98, and 0.99,
corresponding to residual energies of at most 10\%, 5\%, 2\%, and 1\% of its fitting-set
variance. Each target is applied separately. We choose the first saved checkpoint whose
fitting EV reaches the target. Native coordinate losses, labels, and held-out scores do
not choose the checkpoint. The thresholds were specified for this retrospective analysis
after the original study's outcomes were known, not as part of its confirmation protocol.

The recording schedule limits the comparison. A target met at update 100 could have been
met earlier. A target first observed at update 2,000 might have been met between saved
observations. Curves can also be nonmonotonic. We therefore report first \emph{saved} hits,
not interpolated crossing times. Targets with no saved hit remain in the record, rather
than being assigned zero effect or excluded from reach counts. Paired differences use
only seeds in which both methods have an observed hit, with the denominator reported.
The original five seeds are reused across targets and metrics, so these rows are not
independent replications.

\begin{figure}[!ht]
  \centering
  \includegraphics[width=\linewidth]{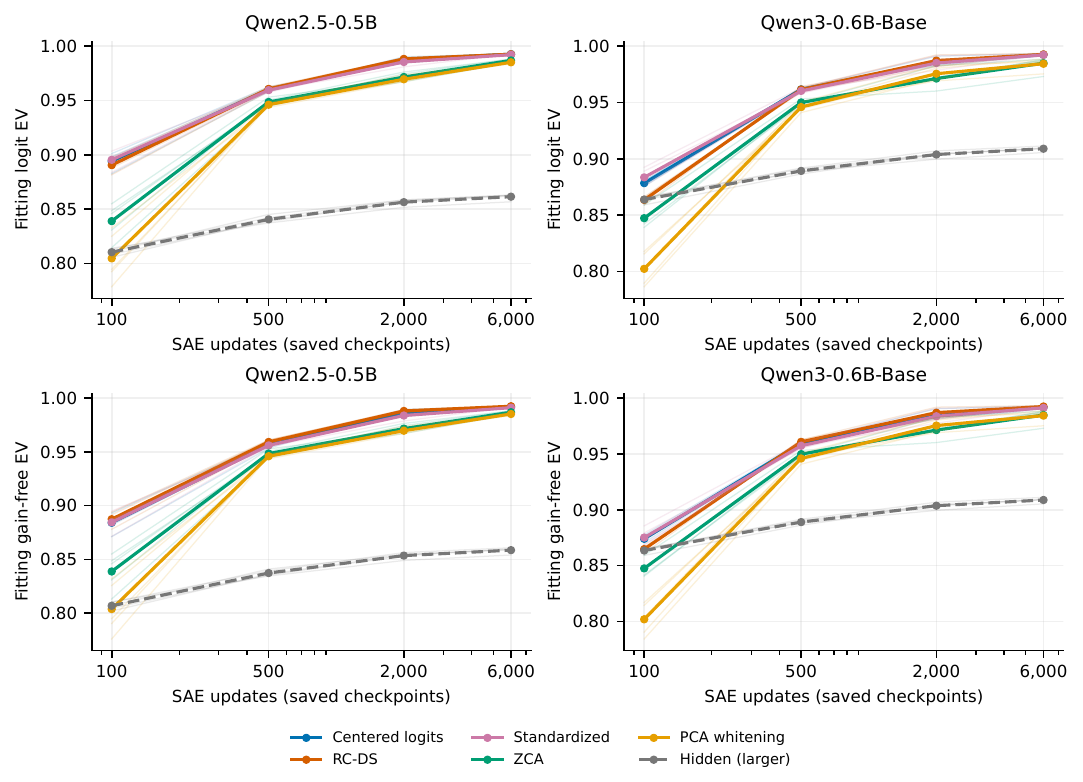}
  \caption{Common fitting-quality curves for every method at $K=4$. Upper panels score
  logit reconstruction and lower panels score gain-free reconstruction. Thin lines show
  individual seeds and thick lines show their means. Only the four saved checkpoints are
  measured. Joining them does not supply intermediate observations. The hidden-state
  reference has more parameters, so an update is not equal compute across all methods.}
  \label{fig:common-quality}
\end{figure}

\begin{table}[!ht]
\centering\small
\caption{Logit fitting-quality thresholds at $K=4$. Hits count RC-DS/logit fits out of five seeds. $n$ counts seeds for which both methods reach the target at a saved checkpoint. Differences are RC-DS minus logits over those pairs only. $\Delta T$ is updates, $\Delta F$ is attained fitting EV, and $\Delta E$ is held-out EV in the same metric. AUC is held-out single-feature macro AUROC. Nonzero $\Delta F$ shows that threshold matching is not exact loss matching.}
\label{tab:quality-logit_ev}
\begin{tabular}{llrrrrrr}
\toprule
Model & Target & Hits & $n$ & $\Delta T$ & $\Delta F$ & $\Delta E$ & $\Delta$AUC \\
\midrule
Qwen2.5 & 0.90 & 5/5 & 5 & +80 & +0.0127 & +0.0135 & +0.0122 \\
Qwen2.5 & 0.95 & 5/5 & 5 & +0 & +0.0004 & +0.0014 & +0.0029 \\
Qwen2.5 & 0.98 & 5/5 & 5 & +0 & +0.0007 & +0.0017 & -0.0014 \\
Qwen2.5 & 0.99 & 5/5 & 5 & +800 & +0.0004 & +0.0012 & +0.0036 \\
Qwen3 & 0.90 & 5/5 & 5 & +0 & -0.0002 & -0.0011 & +0.0055 \\
Qwen3 & 0.95 & 5/5 & 5 & +0 & -0.0002 & -0.0011 & +0.0055 \\
Qwen3 & 0.98 & 5/5 & 5 & +0 & +0.0010 & +0.0004 & +0.0003 \\
Qwen3 & 0.99 & 5/5 & 5 & +0 & +0.0004 & +0.0003 & -0.0038 \\
\bottomrule
\end{tabular}
\end{table}

\begin{table}[!ht]
\centering\small
\caption{Gain-free fitting-quality thresholds at $K=4$. Hits count RC-DS/logit fits out of five seeds. $n$ counts seeds for which both methods reach the target at a saved checkpoint. Differences are RC-DS minus logits over those pairs only. $\Delta T$ is updates, $\Delta F$ is attained fitting EV, and $\Delta E$ is held-out EV in the same metric. AUC is held-out single-feature macro AUROC. Nonzero $\Delta F$ shows that threshold matching is not exact loss matching.}
\label{tab:quality-gain_free_ev}
\begin{tabular}{llrrrrrr}
\toprule
Model & Target & Hits & $n$ & $\Delta T$ & $\Delta F$ & $\Delta E$ & $\Delta$AUC \\
\midrule
Qwen2.5 & 0.90 & 5/5 & 5 & +0 & +0.0025 & +0.0037 & +0.0029 \\
Qwen2.5 & 0.95 & 5/5 & 5 & +0 & +0.0025 & +0.0037 & +0.0029 \\
Qwen2.5 & 0.98 & 5/5 & 5 & +0 & +0.0018 & +0.0027 & -0.0014 \\
Qwen2.5 & 0.99 & 5/5 & 5 & +0 & +0.0007 & +0.0014 & +0.0024 \\
Qwen3 & 0.90 & 5/5 & 5 & +0 & +0.0013 & +0.0002 & +0.0055 \\
Qwen3 & 0.95 & 5/5 & 5 & +0 & +0.0013 & +0.0002 & +0.0055 \\
Qwen3 & 0.98 & 5/5 & 5 & +0 & +0.0017 & +0.0011 & +0.0003 \\
Qwen3 & 0.99 & 5/5 & 5 & +0 & +0.0010 & +0.0008 & -0.0038 \\
\bottomrule
\end{tabular}
\end{table}

The tables focus on the original primary $K=4$ contrast, but the machine-readable record
retains all 1,440 fit--metric--target rows across all methods and sparsities. It includes
attained fitting EV, overshoot, the previous saved observation, missed targets, training-only
seconds, updates, examples seen, and estimated dense matrix-multiplication work. At $K=4$,
all ten RC-DS fits and all ten logit fits reach every target. Gain-free targets select the
same saved checkpoint for both methods in every seed. For logit targets, one Qwen2.5 seed
selects RC-DS at 500 rather than 100 updates for the 0.90 target, and another selects RC-DS
at 6,000 rather than 2,000 for the 0.99 target. All other logit-target pairs select the same
saved update. This recording schedule supplies no RC-DS update-count advantage. Original
training times are retained, but do not establish an end-to-end speedup.

For the new descriptive label-probe analysis, one feature and its sign per class are
selected by AUROC on the original 3,000 fitting examples at each checkpoint. Labels do
not enter SAE training or fitting-quality selection. We hold the probes fixed and measure
class-macro AUROC on the original 360 disjoint prompts. Thus, feature selection reuses SAE
fitting data, and evaluation reuses the previously opened prompt panel. This is distinct
from the full-vocabulary study's separate 600-text feature-selection split. It measures
single-feature label detection, not causal steering or ground-truth recovery. The signs
of the RC-DS--logit AUROC differences vary across models and targets, and checkpoint
overshoot remains a possible source of differences.

\begin{table}[!ht]
\centering\small
\caption{Original Qwen2.5 learning-rate search cost at $K=4$, six complete fits per method (three rates, two seeds). Qwen3 reuses these choices. Training-only seconds exclude extraction, preprocessing, evaluation, and serialization.}
\label{tab:quality-tuning}
\begin{tabular}{lrrr}
\toprule
Method & Updates & Training seconds & Dense work ($10^9$ FLOPs) \\
\midrule
Centered logits & 36,000 & 18.21 & 42.47 \\
RC-DS & 36,000 & 18.11 & 42.47 \\
Standardized & 36,000 & 18.21 & 42.47 \\
ZCA & 36,000 & 18.48 & 42.47 \\
PCA whitening & 36,000 & 18.36 & 42.47 \\
Hidden (larger) & 36,000 & 67.19 & 6342 \\
\bottomrule
\end{tabular}
\end{table}

Every method and sparsity used six full Qwen2.5 tuning fits before the confirmation fits.
Qwen3 reused the selected rates without an additional search. Early checkpoints still
inherit choices made using the 6,000-update tuning endpoints, so their selection cost is
not zero. This analysis does not retune specifically for an early quality target or compare
the best attainable speed of each method.

\FloatBarrier
\subsection{Varying factor correlation and the number of active factors}
\begin{samepage}
The synthetic comparison uses the same independent-encoder trainer with 16 atoms in
eight hidden dimensions, 2,048 fitting and 512 test samples, and 3,000 updates. Factors are
unit-normalized mixtures of independent random directions and a shared unit direction with
square-root weights $\sqrt{1-\rho}$ and $\sqrt{\rho}$. The construction parameter
$\rho\in\{0,0.7\}$ is not a measured pairwise correlation. Codes have one, two, or four active
factors, chosen uniformly with nonzero coefficients sampled uniformly from $[0.5,1.5]$.
The SAE's $K$ matches this generating activity. Readouts have condition numbers one or 100.
All four coordinate methods, both learning rates $0.001$ and $0.003$, and all three seeds
270401--270403 are reported, for 288 fits. No rate is selected on recovery. The full grid,
including condition number one, is retained in the internal research record. Factors are generated in the
gain-free space, so this test supplements rather than replaces the generating-coordinate
control.
The recovery score is also harder to interpret when the true directions are highly correlated:
one learned direction can be similar to several true factors. We therefore compare methods
within each correlation setting, not raw scores across correlation settings. The score uses
signed cosine similarity, not its absolute value.
\par
\end{samepage}

\begin{figure}[!ht]
  \centering
  \includegraphics[width=\linewidth]{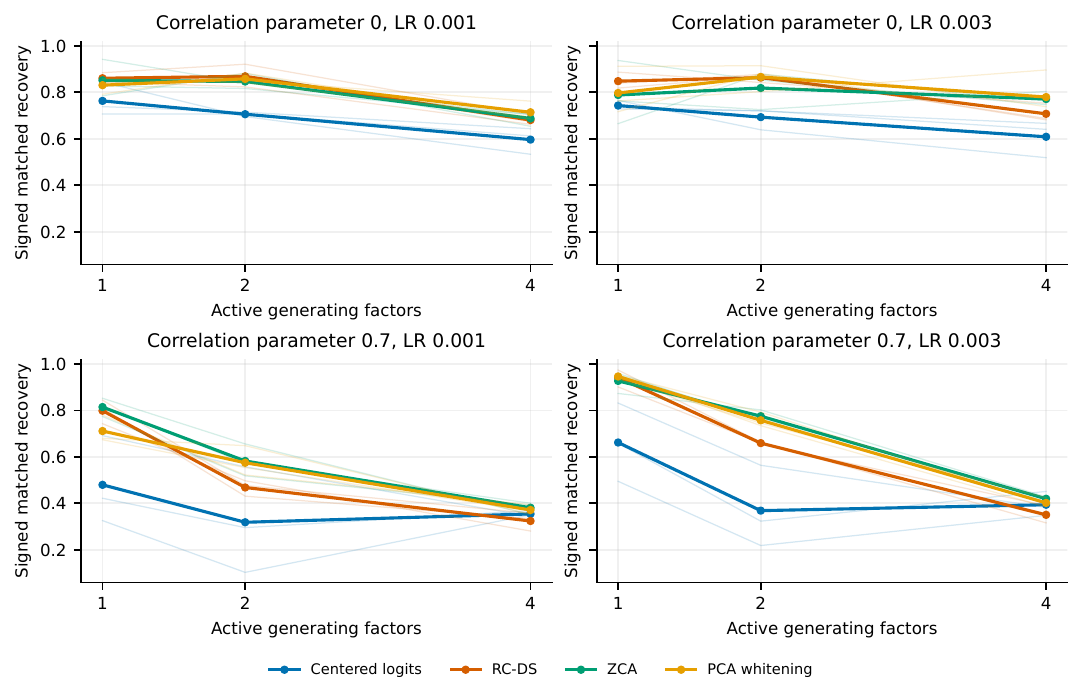}
  \caption{Factor recovery at condition number 100 as correlation and simultaneous activity
  change. Both learning rates are shown. Thin lines show individual seeds and thick
  lines show means. Recovery is signed cosine similarity after mapping decoder directions
  back to the known generating space.}
\end{figure}

\FloatBarrier
\section{Qwen Full-Vocabulary Reconstruction}
\label{app:qwen-full-readout}

\subsection{Readout, data, and optimization}

The centered Qwen2.5-0.5B vocabulary matrix has shape $151{,}936\times896$, retained rank
896, and condition number 22.6. The spectrum uses float64 Gram eigenvalues and relative
rank cutoff $10^{-6}$. Tied embeddings are verified against the output head. Pinned model
revisions, weight checksums, coordinate factors, and the prepared-state checksum are retained
in the internal full-readout records.

The comparison uses 3,000 cached Emotion fitting states and splits the previously used 480
development states into 240 tuning and 240 scoring rows with seed 267000. Fitting means,
coordinate scales, and covariance statistics use fitting rows only. A token-sequence audit
finds one fitting--tuning and one fitting--scoring duplicate, with no tuning--scoring overlap.
There are 2,995 unique fitting sequences. The recorded data were not changed after this
discovery. These are reused development data with known duplicates, not a new deduplicated test set.

All eight methods use 896-dimensional inputs, 1,792 decoder atoms, Top-$K$ 16, independently
updated encoders, paired data-row initialization, and the same 400-epoch positive-floor
learning-rate schedule. SAE seeds are 267001 and 267002. Every arm receives rates 0.001
and 0.003. Mean tuning logit EV selects its rate before scoring. Dictionary expansion two
and Top-$K$ 16 were fixed for Qwen after a separate development search, not selected on
Qwen's scoring rows. The full search record is retained internally.

The compact RC-DS input is $q=V^\top(h-\mu)$, and the compact-logit input is $\Sigma q$. These
factorizations exactly preserve ambient RC-DS and logit reconstruction norms, but do not
claim to reproduce unconstrained vocabulary-width Adam updates. At full rank, the
gain-free coordinates are an orthogonal rotation of hidden states. Ordinary hidden-state
SAEs are therefore an essential control. Standardization and PCA/ZCA whitening act in
the intrinsic basis, with fitting-only relative variance floors of $10^{-6}$.
Whitened SAEs minimize MSE in whitened coordinates. They do not test whitening the encoder
input while retaining an unwhitened output loss. The inverse-loss arm changes only the weighting
of residuals in compact-logit coordinates, not the complete encoder/decoder wrapper.

\subsection{Reconstruction and training checks for every method}

\begin{table}[!ht]
  \centering
  \small
  \caption{Qwen full-readout scoring reconstruction. Means use both SAE seeds at each arm's
  tuning-selected learning rate. The metrics use the same predictions in different spaces.
  Neither is known-factor recovery.}
  \label{tab:app-qwen-full-readout}
  \begin{tabular}{lrr}
    \toprule
    Training method & Logit EV & Gain-free EV \\
    \midrule
    RC-DS & 0.7741 & 0.7928 \\
    Half gain & 0.7816 & 0.7922 \\
    Centered logits & 0.7835 & 0.7882 \\
    Hidden states & 0.7725 & 0.7908 \\
    Standardized & 0.7765 & 0.7802 \\
    PCA whitening & 0.5138 & 0.5021 \\
    ZCA whitening & 0.5279 & 0.5204 \\
    Inverse-gain loss & 0.7784 & 0.7886 \\
    \bottomrule
  \end{tabular}
\end{table}

Centered logits are the baseline selected using tuning data. RC-DS minus centered-logit EV
is $-0.009332$. RC-DS
logit EV is 0.7738 and 0.7745 across the two seeds, compared with 0.7820 and
0.7849 for centered logits. Its mean gain-free advantage is 0.004571 over the logit
SAE and 0.002022 over the hidden-state SAE. These are descriptive two-seed differences,
not confidence intervals over models.

The comparison comprises 32 complete fits. Mean effective activity is exactly 16 for
every selected arm. Six of eight selected arms exceed the 1\% final-window loss-change
diagnostic, comparing mean fitting loss over epochs 301--350 and 351--400.
RC-DS scoring logit EV changes by $-0.000586$ between the 200- and 400-epoch snapshots
under the fixed schedule. Scoring snapshots do not choose the stopping epoch.
Equal budgets therefore do not establish equally converged models, and the weak whitening
outcomes do not establish superiority over all whitening procedures.

\FloatBarrier
\section{Steering Results and Controls}
\label{app:causal-boundary}

The game studies select groups of decoder directions without using labels or rewards. They
compare them with direct logit changes and eight random dictionaries selected in the same way.
The decoder families allow fixed action anchors. Bargaining's family was developed after
inspecting the policies, so its intervention evidence is post hoc.
Bargaining's intervention is nearly a direct change to the target-action logit: the component
orthogonal to that direction is only
0.032\%--7.47\% of its norm. Goofspiel's selected family increases expected bid on prize cards
5--7, but its aggregate gains 0.0319, 0.0613, and 0.0536 are smaller than the best random
gains 0.0523, 0.0692, and 0.0954 across $\alpha=0,1/2,1$. Decoder-family steering is therefore
not evidence of a uniquely useful learned feature. A stricter, naturally active contextual bargaining
study finds that an optimized fixed direction misses 51.31\% of the oracle's achievable
improvement. Yet the best learned intervention beats the strongest control by only 2.44\%
of the oracle's improvement, below the predeclared 5\% requirement.
Reserved contextual bargaining policies are not opened.

The constrained study uses Qwen2.5-0.5B \citep{qwen2024qwen25} on the Emotion dataset
\citep{saravia2018carer}, with six label-token logits,
3,000 fitting examples, 480 development examples, 360 balanced test examples, and five
predeclared SAE seeds. The 60-epoch anchored trainer and intervention protocol are fixed
before confirmation.
Naturally active non-anchor features, an optimized fixed direction, and eight selection-matched
random dictionaries share a pointwise KL budget of 0.05. The oracle uses the known row label.
This is controlled steering, not independent label prediction. Mean excess is learned gain minus
the strongest control, normalized by oracle gain.

\begin{table}[!ht]
  \centering
  \small
  \caption{Frozen Qwen Emotion test. The seed-bootstrap interval and practical criterion count
  retain every seed. A positive learned margin does not isolate the coordinate choice.}
  \label{tab:app-causal-boundary}
  \begin{tabular}{lrrr}
    \toprule
    Model & Mean excess & 95\% seed interval & Seeds above 5\% \\
    \midrule
    Qwen2.5-0.5B & $+7.73\%$ & $[+4.37\%,+11.18\%]$ & 4/5 \\
    \bottomrule
  \end{tabular}
\end{table}

All five Qwen seed effects are positive, with four above 5\%. Qwen passes the full rule of
at least four positive seeds, at least three above 5\%, and
all correctness checks. This study includes fixed anchor directions, unlike the later
coordinate comparison in which every atom is learned. That comparison's RC-DS minus
centered-logit steering difference at the training cap is $-4.08\%$ of
oracle gain, with seed interval $[-10.73\%,+0.82\%]$ and prompt interval
$[-5.42\%,-2.77\%]$. A prompt interval for this fixed seed panel does not estimate variation
over new SAE training seeds.

Two frozen Qwen sentiment-generation studies use independent DistilBERT SST-2 scoring
\citep{sanh2019distilbert}. Both use the same released SST-2-finetuned checkpoint, pinned
to a fixed revision.\footnote{Checkpoint and revision:
\url{https://huggingface.co/distilbert/distilbert-base-uncased-finetuned-sst-2-english/tree/714eb0fa89d2f80546fda750413ed43d93601a13}.}
The first fits SAEs on IMDb states \citep{maas2011imdb}, evaluates
15 neutral prefixes adapted from PPLM \citep{dathathri2020pplm}, and has learned-minus-random
target probability $+0.0138$ with prefix interval $[-0.0220,+0.0504]$. The second fits on
generated states and uses disjoint OpenWebText prompts from DExperts \citep{liu2021dexperts}.
Its mean is $-0.0025$ with prompt interval $[-0.0255,+0.0207]$. Both use five SAE seeds,
paired continuations, and a per-token KL cap of 0.01, and both fail their complete confirmation
criteria. Later sequence-aware development has a positive mean $+0.0370$ but interval
$[-0.0255,+0.1016]$, also insufficient to establish a reliable learned advantage. These test steering during generation,
not reconstruction across full-vocabulary coordinates.

Feature activity diverges during generation despite matching on fitting data. In the first
study, the selected learned family is active on 3.53\% of generated states versus 26.35\%
for its random control. In the generated-state follow-up, four of ten seed--polarity
activity ratios miss the declared range. Directions with the intended immediate token-axis
sign can also lower externally scored positive sentiment after generation. Immediate logit
effects and sequence-level sentiment are separate outcomes.

\FloatBarrier
\section{Experimental Setup and Verification}
\label{app:reproducibility}

\subsection{Software and computation}

The studies use Python and PyTorch for policy and SAE training, OpenSpiel for the game
environments, and Hugging Face Transformers for the pretrained language models and tokenizers.
Game policies are trained before their representations are collected. The pretrained language
models remain frozen. Their hidden states and readout factors are cached so that coordinate
methods can be compared on the same examples without repeating model inference for every SAE fit.

The 540-fit full-vocabulary study runs SAE training with three concurrent CPU workers and
one thread per fit. Development and confirmation training take 2.66 and 5.62 wall-clock hours,
respectively, excluding representation extraction and final scoring. These timings describe
that study, not the total cost of the paper. Appendix~\ref{app:full-readout-usefulness} gives
the model dimensions, data sizes, update schedule, and per-fit compute totals.

\subsection{Data partitions and matched training}

Each study distinguishes examples used to fit the SAE, select methods or features, and
evaluate the selected result. Centering means, normalization scales, and covariance transforms
are estimated from fitting examples. Model and dataset revisions, numerical rank tolerances,
seeds, and split rules are specified by the study's protocol. Coordinate comparisons match
SAE capacity and training budgets within a study. Paired initialization and minibatch schedules
are used where specified, and learning-rate selection follows the stated rule for each study.
Table~\ref{tab:study-overview} at the start of the appendices summarizes the distinct data,
budgets, and questions, with links to each study's detailed settings.

The attribution and training studies use previously opened development data. The training
audit selects no recipe and evaluates no fresh interventions. The Qwen full-readout comparison
also uses previously opened prompts. These analyses are distinct from the separately frozen
game and text evaluations. The manuscript's case studies were selected after broader
exploration, not as a preregistered model survey.

\subsection{Result verification}

Saved records include training histories, checkpoints, selection decisions, and checksums
linking results to their protocols and prepared data. Paper tables and figures are generated
from these saved fits without retraining. For example, the Qwen full-readout comparison retains
all 32 fits and the results for all eight methods. The 540-fit full-vocabulary study replays
1,080 saved checkpoints and checks 585 semantic score rows, including feature selection,
tie-aware AUROC, text separation, selected learning rates, and forecast timestamps.

The internal experiment ledger retains the broader research history,
including outcomes outside the manuscript's scope. The appendices provide the study setup,
mathematical derivations, evaluation procedures, and supporting results.

\section{Observable Recovery Prediction and Semantic Probes}
\label{app:observable-recovery}

\subsection{An observable diagnostic with no true-factor inputs}

The earlier cosine-recovery score measures signed agreement between decoder directions and
the generating directions. Highly correlated true factors can make this score difficult to
interpret. We therefore also measure support F1. A one-to-one feature assignment maximizes
activation-support overlap on fitting examples only. Holding that assignment fixed, we compute
micro F1 between positive SAE features and the true active factors on separate evaluation rows.
True codes are used for assignment and evaluation, never for SAE training.

The predictor receives fitting hidden activations, the centered readout, the configured $K$,
and dictionary width. It has no true-factor, coefficient, label, or evaluation-row input.
Let $u_i$ be each mean-centered fitting example normalized to unit length. We measure
\begin{equation}
 a=\frac{1}{n(n-1)}\sum_{i\ne j}(u_i^\top u_j)^2.
\end{equation}
This is overlap between fitting examples, not direct observation of latent-factor overlap.
Let $c=a_{\rm logits}-a_{\rm RC}$, $\bar a$ be the mean of those two overlaps,
and $b$ the difference in log covariance condition numbers. The retained regression uses
$c$, $cK/m$, $c\bar a$, and $b$, where $m$ is dictionary width. It has zero intercept and
a numerical ridge of $10^{-8}$. Its coefficients fitted to all development rows, rounded for display, are
$(3.634,-4.847,-6.283,0.002888)$.

Six autoresearch trials use only previously observed outcomes. They reduce grouped
development MAE from 0.202 for a zero-effect prediction to 0.0538. All seeds and condition
numbers sharing a generating-coordinate, correlation, and sparsity setting remain together
in each fold. There are 68 nonidentity rows in eleven groups. The target of 0.05 is not reached.
Previously observed confirmation rows are explicitly development data for this new search.

\subsection{Fresh controlled validation}

After freezing the predictor, we evaluate twelve new settings with three fresh seeds each.
All 36 forecasts precede their SAE fits. Eight settings cross generating-coordinate exponent
$\beta\in\{0,1\}$, correlation parameter $\rho\in\{0,0.7\}$, and $K\in\{2,4\}$ at
readout condition number 100. Four further settings use condition number 30, $\rho=0.35$,
and $K=3$. They include intermediate exponents 0.25 and 0.75, and $\beta=0,1$ with noise.

Each setting has eight dimensions, sixteen factors, 2,048 fitting rows, and 512 evaluation
rows. Factors are drawn in a base coordinate system and transformed by $\Sigma^{-\beta}$
without a second normalization. Active positive coefficients are uniform on $[0.5,1.5]$.
Noise is isotropic before that transformation, with RMS 0.05 times the clean centered fitting
RMS. Signal and noise undergo the same map, preserving information equivalence between arms.

Centered-logit, RC-DS, standardization, and ZCA SAEs each receive rates 0.001 and 0.003.
There are 288 fits with 6,000 updates each, batch size 256, and independently optimized encoder
and decoder weights. The decay schedule remains anchored to the first 3,000 updates.
Each arm chooses its rate by final native fitting MSE only. Both rates are retained.

The frozen predictor gets the direction right in all 36 pairs. All absolute observed effects
exceed the declared 0.03 F1 band. MAE is 0.0526, versus 0.241 for zero-effect prediction.
A bootstrap over the twelve whole settings gives $[0.0399,0.0685]$ for MAE. The 0.05 target
remains missed. These are twelve settings with three seeds, not 36 independent real networks.
ZCA has higher support F1 than RC-DS in 34 of the 36 pairs.

At $\rho=0$, $K=2$, and condition number 100, mean support F1 is 0.7878 for RC-DS versus
0.3783 for logits when $\beta=0$. At $\beta=1$, the ordering reverses to 0.3418 versus
0.7878. The same condition number can therefore accompany a large benefit or a large cost
from removing the gains. The frozen observable diagnostic predicts these reversals within
the tested family. It is not a distribution-independent recovery theorem.

\subsection{Semantic usefulness on the two constrained Qwen heads}

The semantic comparison reuses pinned fitting activations from Qwen2.5-0.5B and Qwen3-0.6B-Base, with
3,000 fitting texts and six-label readouts. It trains four coordinate methods, two sparsities,
three seeds, and two rates, for 96 fits. SAEs have 64 atoms and train for 6,000 updates.
For each class, one feature and its sign are selected by fitting AUROC. The fixed feature
is then evaluated on disjoint texts. Ties contribute one half in the exact rank-based AUROC.

The evaluation panel is balanced with 80 texts per class. It comes from the larger unsplit
Emotion pool, excluding all 20,000 official split texts and duplicate normalized or truncated
token identities. It is not an official benchmark test split, and pretraining contamination
cannot be excluded. The initial requested pool had too few unused examples in one class.
That attempt stopped before new activations or evaluation scores were obtained. A separate
sampling amendment was recorded, and all fitting MSE values were identical on the retry.
The failed attempt and amendment remain in the audit record.

\begin{table}[!ht]
  \centering\small
  \caption{RC-DS minus centered-logit semantic AUROC on 480 new texts. Seed ranges are
  conditional on each fixed checkpoint and text panel, not confidence intervals over models.}
  \begin{tabular}{llrr}
    \toprule
    Model & $K$ & Mean difference & Seed range \\
    \midrule
    Qwen2.5 & 1 & $+0.00095$ & $[-0.00455,+0.00688]$ \\
    Qwen2.5 & 4 & $-0.00603$ & $[-0.00834,-0.00170]$ \\
    Qwen3 & 1 & $+0.00289$ & $[+0.00162,+0.00357]$ \\
    Qwen3 & 4 & $+0.00142$ & $[-0.00519,+0.01019]$ \\
    \bottomrule
  \end{tabular}
\end{table}

\subsection{Internal verification}

The private research record contains the training and analysis code with saved outputs. The 288-fit
observable validation has 576 replayed training checkpoints. The semantic study has 192
replayed checkpoints and 104 replayed semantic-score rows, including random-initialization
and raw-coordinate controls. The six-trial predictor search and unsuccessful hypotheses
are preserved separately, together with source hashes.

\section{Additional Reconstruction Studies}
\label{app:prior-reconstruction}

The synthetic study measures known-factor recovery. Real-model generating factors are unknown,
so we instead test reconstruction: can RC-DS reproduce the same logits more accurately, and
does any advantage remain when baselines receive more training and separate tuning?

\subsection{Fixed-budget improvements in neural policies}

Actor--critic policies train in OpenSpiel bargaining and seven-card imperfect-information
Goofspiel for 3,000 and 5,000 episodes, respectively. Both use 192-atom, Top-$K$ 8 SAEs
trained for 15 and 125 epochs, respectively. Each environment contributes three evaluation
policies, with separate episodes for SAE training and evaluation. Bargaining's policies had
already been inspected during development, so their evaluation is post-hoc. Goofspiel uses
new held-out policy seeds.

\begin{table}[!ht]
  \centering
  \small
  \caption{Shared centered-logit EV. Game rows average three policies under their fixed budgets.
  Qwen rows average five paired SAE seeds on reused development data.}
  \label{tab:reconstruction}
  \begin{tabular}{lrrr}
    \toprule
    Setting & Centered logits & RC-DS & Difference \\
    \midrule
    Bargaining & 0.8802 & 0.9144 & $+0.0342$ \\
    Goofspiel & 0.7654 & 0.8637 & $+0.0983$ \\
    Qwen label head, 60 epochs & 0.6066 & 0.6264 & $+0.0198$ \\
    Qwen label head, 1,000 epochs & 0.9639 & 0.9655 & $+0.0016$ \\
    \bottomrule
  \end{tabular}
\end{table}

\begin{samepage}
RC-DS reconstruction is higher for all three bargaining policies and two of three Goofspiel
policies. The Goofspiel differences are $-0.0045,+0.1085,+0.1909$, so the average hides
variation between policies. Reweighting only the loss performs substantially worse in both
games. Unlike the exact wrapper, it still feeds scaled inputs to the encoder, with scaling
that depends on the allowed actions. These results cover few policies and fixed training
budgets. They do not show that RC-DS beats fully converged baselines.
Protocols and per-policy results are in Appendix~\ref{app:policy-reconstruction}.
\par
\end{samepage}

\subsection{Longer training nearly closes the Qwen gap}
\label{sec:coordinate-training}

For Qwen, we restrict output to six single-token Emotion labels. A comparison with all 64
dictionary atoms learned uses 3,000 training states, 480 previously used development states,
five paired SAE seeds, and Top-$K$ 4. Hidden-state, centered-logit, RC-DS, and ZCA-whitened
SAEs use the same initialization examples and minibatch order. The label head has rank five
and $\kappa=1.425$. Each action-space SAE has 838 parameters, whereas the hidden-state SAE
has 115,648. The latter is a larger reference model, not a parameter-matched baseline.

Extending the same fits from 60 to 1,000 epochs reduces the RC-DS--logit difference from
0.0198 to 0.0016. ZCA reaches 0.9690. None of the action-space fits meets the predeclared
loss-stability rule at the training cap, so the final scores do not establish convergence.
This trainer also copies decoder weights into the encoder after each Adam step. A separate
48-fit comparison finds that letting encoder weights learn independently improves training
MSE in every representation. No setting meets both its accuracy and stability requirements,
so its reserved evaluation data remain unused. We therefore test independent encoder updates
and separate learning-rate selection next. Full procedures, outcomes, and training curves are in
Appendix~\ref{app:coordinate-training}.

\subsection{Separate tuning gives near-ties on new prompts}
\label{sec:fresh-tradeoffs}

We compare centered logits, RC-DS, standardization, PCA whitening, ZCA whitening, and the
larger hidden-state reference. Each SAE has 64 atoms and trains for 6,000 Adam updates with
batch size 256 and independent encoder and decoder updates. On Qwen2.5-0.5B
\citep{qwen2024qwen25}, each method and $K\in\{1,2,4\}$ gets three learning rates and two
tuning seeds. We select rates separately by mean final logit EV on the reused 480 development
prompts. Five new SAE seeds then train each selected setting on Qwen2.5 and Qwen3-0.6B-Base
\citep{qwen2025qwen3}. Qwen3 reuses Qwen2.5's selected rates without retuning. It tests whether
those settings transfer, not the best separately tuned result on Qwen3.

All 180 trained SAEs are saved before we evaluate 360 new Emotion prompts, balanced across
labels. These exclude SAE training texts and the official validation and test texts, with
duplicate checks on both text and token sequences. The predeclared primary comparison is
RC-DS minus centered-logit EV at $K=4$ after 6,000 updates. Other sparsity levels and training
times are secondary comparisons. Both heads have rank five and nearly equal gains, with
condition numbers 1.425 and 1.353. This tests two small label heads, not their full vocabularies.

\begin{table}[!ht]
  \centering\small
  \caption{RC-DS and centered-logit SAEs are nearly tied on new prompts at $K=4$.
  Intervals resample the five paired SAE seeds for each fixed model and the same 360 prompts.
  Both intervals include zero. They do not measure uncertainty across language models.}
  \label{tab:fresh-primary}
  \begin{tabular}{lrrrl}
    \toprule
    Checkpoint & Logit EV & RC-DS EV & Difference & 95\% seed interval \\
    \midrule
    Qwen2.5-0.5B & 0.9920 & 0.9926 & $+0.00058$ & $[-0.00011,+0.00128]$ \\
    Qwen3-0.6B-Base & 0.9918 & 0.9919 & $+0.00008$ & $[-0.00106,+0.00141]$ \\
    \bottomrule
  \end{tabular}
\end{table}

\begin{figure}[!ht]
  \centering
  \includegraphics[width=\linewidth]{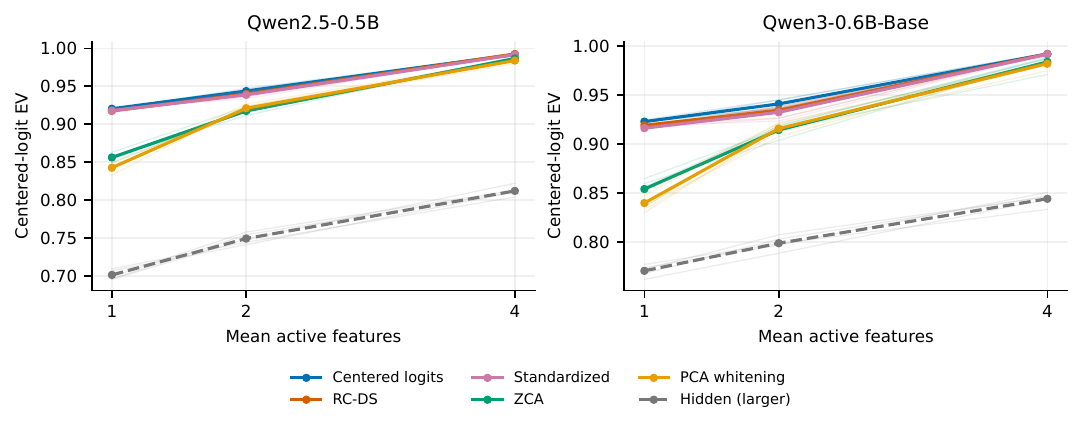}
  \caption{Reconstruction on new prompts with one, two, or four active features after 6,000 updates.
  Thin lines show all five seeds and thick lines show means. The larger hidden-state arm is a
  reference, not a parameter-matched baseline. Rates are tuned per method and $K$ on Qwen2.5 and
  transferred unchanged to Qwen3.}
  \label{fig:fresh-sparsity}
\end{figure}

The primary comparison does not establish a reliable RC-DS advantage
(Table~\ref{tab:fresh-primary}). At one or two active features, centered-logit training gives
higher mean logit EV on both models. Whitening also gives lower logit EV under these settings,
but this does not show that its features are less interpretable. Figure~\ref{fig:fresh-sparsity}
shows all methods and sparsity levels. Training loss still changes by more than 1\% near the
end in 31 of 180 fits, including three of five primary logit fits on each model. Equal tuning
and training budgets therefore do not establish equal convergence. Appendix~\ref{app:fresh-tradeoffs}
reports the loss check, gain-free EV, training-time plots, and full protocol.

\subsection{Full-vocabulary results depend on the scoring metric}
\label{sec:qwen-full-readout}

We next keep Qwen2.5's entire output vocabulary instead of six labels. Its centered head has
rank 896 and condition number 22.6, compared with 1.425 for the label head. SAEs train in
896-dimensional coordinates with 1,792 atoms, Top-$K$ 16, two seeds, and 400 epochs. Each of
eight coordinate or loss choices gets the same two learning rates, selected by tuning logit EV.
The 3,000 training states and 240/240 tuning/evaluation split come from previously used Emotion
prompts. These are development results, not confirmation on new data.

On evaluation rows, RC-DS logit EV is 0.7741 versus 0.7835 for centered-logit training,
a difference of $-0.0093$. Gain-free EV instead favors RC-DS, 0.7928 versus 0.7882, but the
ordinary hidden-state SAE is close at 0.7908. Both SAE seeds have the same sign of the logit
difference. Thus, the higher condition number does not produce a common-logit reconstruction
advantage in this comparison. The two metrics weight the same residuals differently
(Equation~\ref{eq:residual-metrics}). Neither measures recovery of known sparse factors.

Six of eight selected methods exceed the 1\% late-training loss-change threshold, so equal
budgets do not establish equal convergence. Comparing two readouts from one model also cannot
show that condition number predicts benefit across models. Appendix~\ref{app:qwen-full-readout}
reports all methods, the reused data and duplicate checks, and the training procedure.

\FloatBarrier
\section{Full-Readout Semantic Usefulness}
\label{app:full-readout-usefulness}

This study tests whether fitting-data geometry predicts a change in label detection, not
whether an SAE recovers unknown generating factors. It uses full pretrained vocabulary
readouts rather than fitted task heads. Qwen2.5-0.5B supplies development data
\citep{qwen2024qwen25}. Qwen3-0.6B-Base and SmolLM2-360M are reserved for confirmation
\citep{qwen2025qwen3,benallal2025smollm2}. No language-model weights are updated.
The retained ranks are 896, 1,024, and 960, respectively. The corresponding condition
numbers are 22.6, 14.4, and 79.4.

\subsection{Texts, coordinates, and training}

We use Emotion, AG News, and IMDb \citep{saravia2018carer,zhang2015character,maas2011imdb}.
Each task has 2,400 SAE-fitting texts, 600 feature-selection texts, and 1,200 scoring texts,
balanced across its six, four, or two classes. These are fixed partitions from the published
training pools, not official benchmark test scores. We exclude duplicate normalized texts
and identical truncated token sequences across roles and tasks for all three tokenizers.
Emotion also excludes all 20,000 official-split texts and the earlier 480-text semantic panel.
The same text partitions are used across models. This matching does not establish independence
from language-model pretraining data.

Inputs are raw text without task instructions or label prefixes. We truncate to 128 tokens and
extract the final nonpadding hidden state before the original vocabulary readout. Centering
means and coordinate transforms use fitting states only. We compare centered logits, RC-DS,
ordinary hidden coordinates, per-singular-coordinate standardization, and ZCA whitening.
The logit and RC-DS arms use compact singular coordinates. These preserve reconstruction
norms and the retained information, but a vocabulary-width SAE can follow a different Adam
training trajectory. Standardization acts on singular coordinates, not individual vocabulary tokens.
Whitening uses fitting covariance with the protocol's $10^{-6}$ regularization floor.

All arms use 1,024 dictionary atoms and $K\in\{8,32\}$. The dictionary is modestly
overcomplete for Qwen2.5 and SmolLM2 and equal-width for Qwen3. Three paired seeds and two
learning rates, 0.001 and 0.003, give 180 development and 360 confirmation fits. Each fit
uses 6,000 updates with batches of 128, independently updated encoder and decoder weights,
unit decoder directions, and global fitting-RMS normalization. The learning rate decays to
one tenth of its initial value by 80\% of the training budget. Paired seeds use the same
sampled fitting rows for initialization and the same minibatch schedule. We preserve initial
weights, complete training histories, and checkpoints at updates 3,000 and 6,000.
For each arm, $K$, and seed, the final native fitting MSE selects the learning rate.
No semantic labels or scoring outcomes are used for this selection.

\subsection{Label detection and frozen forecasts}

For each class, we choose one feature and its orientation on the separate 600-text selection
split. On the 1,200 scoring texts, we compute exact tie-aware one-versus-rest AUROC, then
average equally over classes. Each arm receives the same label-selection procedure.
Untrained paired-initialization SAEs and raw-coordinate probes undergo the same procedure.
The primary contrast is the trained RC-DS AUROC minus trained centered-logit AUROC, averaged
over the three paired seeds for each model, task, and $K$.

The development search permits at most six small predictors and leaves out one entire
Qwen2.5 task at a time. It ends after three trials, whose errors relative to the fold-training
mean baseline are 1.681, 1.294, and 0.7326. The last trial meets the declared 0.8 target.
However, its absolute development error is 0.0238 AUROC, worse than the zero-effect
baseline's 0.0227. Thus, meeting that search target is not evidence that useful prediction
has been achieved. There are only three development task groups.

The retained predictor is ridge regression with three inputs: the logit-minus-RC-DS
covariance spectral entropy, the difference in normalized radial fourth moments, and
$\log(K/1024)$. Covariates are standardized using development rows only. Standardized query
values are clipped to $[-2,2]$. With $n$ development rows, ridge penalty $4nI$ regularizes all
coefficients, including the intercept, toward zero effect. Penalizing the intercept makes
zero difference, rather than the development-set mean difference, the default under strong
regularization. This was a deliberate choice given the three development task groups, not
the usual unpenalized-intercept ridge convention. The implementation and every reserved forecast are saved
before any confirmation SAE starts. No confirmation outcome changes this predictor.
The earlier synthetic predictor is also frozen and recorded, but its support-F1 magnitude
is not treated as a calibrated AUROC forecast.

\subsection{Results on the reserved models}

The retained predictor's reserved-model MAE is 0.0169, compared with 0.0135 for the
development-mean constant and 0.00944 for zero effect. Separately, predictor MAE is 0.0183
on Qwen3 and 0.0155 on SmolLM2, versus zero-effect errors of 0.00825 and 0.0106.
Choosing coordinates from its predicted sign incurs mean regret 0.00675 AUROC relative to
choosing the better of RC-DS and logits for each setting after seeing both outcomes.

\begin{figure}[!ht]
  \centering
  \includegraphics[width=\linewidth]{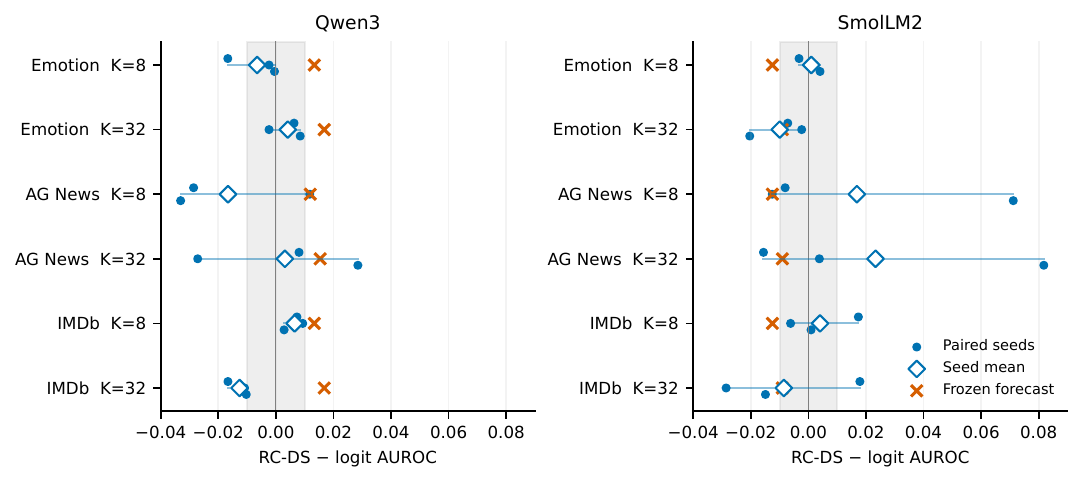}
  \caption{All twelve reserved full-readout settings. Dots show paired seeds, diamonds their
  means, and crosses forecasts frozen before confirmation training. Horizontal lines span
  the three seed values and are not confidence intervals. Gray shading marks the declared
  $\pm0.01$ AUROC band. Each panel is one checkpoint, not six independent models.}
  \label{fig:full-semantic}
\end{figure}

Two setting means exceed $+0.01$, two are below $-0.01$, and eight lie within that band.
The predictor gets all four outside-band directions wrong (Figure~\ref{fig:full-semantic}).
The band was declared before confirmation and is descriptive, not a significance threshold.
Both positive SmolLM2/AG News settings have negative and positive paired seeds. The study
does not support predictable semantic benefits from the tested geometry features.
This finding does not refute fixed-mask equivalence or the controlled synthetic results.

\begin{table}[!ht]
  \centering
  \caption{All trained RC-DS minus centered-logit AUROC differences. Seeds 1--3 are
  730201--730203. Qwen2.5 is development, not an additional reserved-model replication.}
  \label{tab:full-semantic-deltas}
  \begin{tabular}{llrrrrr}
\toprule
Model & Task & $K$ & Mean & Seed 1 & Seed 2 & Seed 3 \\
\midrule
Qwen2.5 & Emotion & 8 & $+0.00047$ & $-0.01294$ & $+0.01430$ & $+0.00004$ \\
Qwen2.5 & Emotion & 32 & $-0.00265$ & $+0.00059$ & $-0.00643$ & $-0.00210$ \\
Qwen2.5 & AG News & 8 & $+0.03900$ & $+0.04816$ & $-0.00584$ & $+0.07467$ \\
Qwen2.5 & AG News & 32 & $+0.04761$ & $+0.05308$ & $+0.04760$ & $+0.04215$ \\
Qwen2.5 & IMDb & 8 & $-0.03294$ & $-0.05240$ & $-0.01532$ & $-0.03109$ \\
Qwen2.5 & IMDb & 32 & $+0.01331$ & $+0.02103$ & $+0.01708$ & $+0.00183$ \\
Qwen3 & Emotion & 8 & $-0.00651$ & $-0.01669$ & $-0.00237$ & $-0.00048$ \\
Qwen3 & Emotion & 32 & $+0.00413$ & $+0.00632$ & $-0.00237$ & $+0.00845$ \\
Qwen3 & AG News & 8 & $-0.01663$ & $-0.02857$ & $+0.01176$ & $-0.03308$ \\
Qwen3 & AG News & 32 & $+0.00313$ & $+0.00805$ & $-0.02715$ & $+0.02849$ \\
Qwen3 & IMDb & 8 & $+0.00650$ & $+0.00732$ & $+0.00933$ & $+0.00286$ \\
Qwen3 & IMDb & 32 & $-0.01261$ & $-0.01665$ & $-0.01088$ & $-0.01029$ \\
SmolLM2 & Emotion & 8 & $+0.00101$ & $-0.00324$ & $+0.00221$ & $+0.00406$ \\
SmolLM2 & Emotion & 32 & $-0.00995$ & $-0.00715$ & $-0.00232$ & $-0.02036$ \\
SmolLM2 & AG News & 8 & $+0.01685$ & $-0.00806$ & $-0.01255$ & $+0.07117$ \\
SmolLM2 & AG News & 32 & $+0.02333$ & $-0.01559$ & $+0.00382$ & $+0.08177$ \\
SmolLM2 & IMDb & 8 & $+0.00404$ & $+0.01737$ & $-0.00618$ & $+0.00093$ \\
SmolLM2 & IMDb & 32 & $-0.00853$ & $+0.01789$ & $-0.02857$ & $-0.01491$ \\
\bottomrule
\end{tabular}

\end{table}

\FloatBarrier
\subsection{All coordinate methods and controls}

Tables~\ref{tab:full-semantic-trained}--\ref{tab:full-semantic-raw} retain every coordinate
method. The hidden and standardized arms are important controls, not only centered logits.
For example, SmolLM2/IMDb at $K=32$ has mean AUROC 0.6078 for hidden coordinates,
0.5507 for logits, and 0.5422 for RC-DS. ZCA's strong synthetic support-recovery results
do not transfer to this semantic score. Several Emotion and IMDb scores are close to chance,
which limits what small differences can establish.
Raw RC-DS coordinate probes have higher AUROC than the corresponding trained RC-DS SAE mean
in all eighteen settings, including development. This is a descriptive comparison, not eighteen
significance tests. Raw RC-DS, compact-logit, and standardized probes are identical here.
Positive per-coordinate scaling and subtraction of a fixed mean preserve example rankings,
so they leave both feature selection and AUROC unchanged.

Table~\ref{tab:raw-vs-trained} places these raw and trained scores side by side. For Qwen2.5
AG News, a raw coordinate scores 0.8493, compared with 0.7053 and 0.8273 for trained RC-DS
features at $K=8$ and $32$. Raw-minus-trained differences range from 0.0002 to 0.1838 AUROC
across the grid. The nine raw model--task scores are each compared with two SAE sparsity
levels. Candidate counts differ: raw probes search the retained rank, while the SAE has
1,024 features. Each trained entry averages three SAE seeds. The table also shows their
observed range, so a raw score exceeding the trained mean is not read as exceeding every
trained run.
For example, Qwen2.5 IMDb at $K=32$ has raw AUROC 0.5650 and trained mean 0.5648,
but trained scores range from 0.5462 to 0.5792. This small mean difference does not
establish a consistent raw-probe advantage across seeds.

\begin{table}[!ht]
  \centering\small
  \caption{Raw singular coordinates versus trained RC-DS features on label detection.
  Each entry is macro-AUROC on the same scoring texts after feature and sign selection on
  separate texts. Trained entries show the mean and [minimum, maximum] over three SAE seeds.
  These are observed ranges, not confidence intervals. The same raw score is compared
  with both sparsity levels. Qwen2.5 supplies development data. The following tables retain
  all other coordinate methods and untrained controls.}
  \label{tab:raw-vs-trained}
  \begin{tabular}{llrrr}
\toprule
Model & Task & Raw & RC-DS, $K=8$ & RC-DS, $K=32$ \\
\midrule
Qwen2.5 & Emotion & 0.5669 & $0.5223\ [0.5128,0.5296]$ & $0.5372\ [0.5349,0.5404]$ \\
Qwen2.5 & AG News & 0.8493 & $0.7053\ [0.6918,0.7130]$ & $0.8273\ [0.8160,0.8351]$ \\
Qwen2.5 & IMDb & 0.5650 & $0.5143\ [0.5045,0.5326]$ & $0.5648\ [0.5462,0.5792]$ \\
Qwen3 & Emotion & 0.5958 & $0.5236\ [0.5152,0.5289]$ & $0.5428\ [0.5392,0.5501]$ \\
Qwen3 & AG News & 0.8471 & $0.6721\ [0.6585,0.6965]$ & $0.7982\ [0.7862,0.8157]$ \\
Qwen3 & IMDb & 0.5564 & $0.5385\ [0.5227,0.5473]$ & $0.5421\ [0.5330,0.5480]$ \\
SmolLM2 & Emotion & 0.5650 & $0.5153\ [0.5117,0.5207]$ & $0.5318\ [0.5292,0.5359]$ \\
SmolLM2 & AG News & 0.8716 & $0.6878\ [0.6579,0.7210]$ & $0.8025\ [0.7873,0.8267]$ \\
SmolLM2 & IMDb & 0.6241 & $0.5016\ [0.4895,0.5129]$ & $0.5422\ [0.5170,0.5809]$ \\
\bottomrule
\end{tabular}

\end{table}

\begin{table}[!ht]
  \centering
  \caption{Trained single-feature macro-AUROC, averaged over three selected SAE seeds.
  Std. denotes per-singular-coordinate standardization. Qwen2.5 is development data.}
  \label{tab:full-semantic-trained}
  \begin{tabular}{llrrrrrr}
\toprule
Model & Task & $K$ & Logits & RC-DS & Hidden & Std. & ZCA \\
\midrule
Qwen2.5 & Emotion & 8 & 0.5219 & 0.5223 & 0.5128 & 0.5226 & 0.5072 \\
Qwen2.5 & Emotion & 32 & 0.5398 & 0.5372 & 0.5408 & 0.5569 & 0.5133 \\
Qwen2.5 & AG News & 8 & 0.6663 & 0.7053 & 0.7138 & 0.6235 & 0.5167 \\
Qwen2.5 & AG News & 32 & 0.7797 & 0.8273 & 0.8313 & 0.7513 & 0.5346 \\
Qwen2.5 & IMDb & 8 & 0.5472 & 0.5143 & 0.5227 & 0.5464 & 0.5008 \\
Qwen2.5 & IMDb & 32 & 0.5515 & 0.5648 & 0.5589 & 0.5673 & 0.5106 \\
Qwen3 & Emotion & 8 & 0.5301 & 0.5236 & 0.5277 & 0.5273 & 0.5044 \\
Qwen3 & Emotion & 32 & 0.5387 & 0.5428 & 0.5482 & 0.5357 & 0.5062 \\
Qwen3 & AG News & 8 & 0.6887 & 0.6721 & 0.6917 & 0.6672 & 0.5147 \\
Qwen3 & AG News & 32 & 0.7951 & 0.7982 & 0.7944 & 0.7810 & 0.5256 \\
Qwen3 & IMDb & 8 & 0.5320 & 0.5385 & 0.5285 & 0.5221 & 0.5100 \\
Qwen3 & IMDb & 32 & 0.5547 & 0.5421 & 0.5347 & 0.5357 & 0.5116 \\
SmolLM2 & Emotion & 8 & 0.5143 & 0.5153 & 0.5124 & 0.5218 & 0.5090 \\
SmolLM2 & Emotion & 32 & 0.5417 & 0.5318 & 0.5319 & 0.5432 & 0.5087 \\
SmolLM2 & AG News & 8 & 0.6709 & 0.6878 & 0.6930 & 0.6628 & 0.5162 \\
SmolLM2 & AG News & 32 & 0.7792 & 0.8025 & 0.7761 & 0.7670 & 0.5243 \\
SmolLM2 & IMDb & 8 & 0.4975 & 0.5016 & 0.5347 & 0.5244 & 0.5165 \\
SmolLM2 & IMDb & 32 & 0.5507 & 0.5422 & 0.6078 & 0.5584 & 0.5116 \\
\bottomrule
\end{tabular}

\end{table}

\begin{table}[!ht]
  \centering
  \caption{Untrained SAE control macro-AUROC. Feature selection uses the same separate texts
  as for trained SAEs. Each value averages the three paired initializations.}
  \label{tab:full-semantic-untrained}
  \begin{tabular}{llrrrrrr}
\toprule
Model & Task & $K$ & Logits & RC-DS & Hidden & Std. & ZCA \\
\midrule
Qwen2.5 & Emotion & 8 & 0.5079 & 0.5062 & 0.5062 & 0.5105 & 0.5040 \\
Qwen2.5 & Emotion & 32 & 0.5253 & 0.5227 & 0.5227 & 0.5297 & 0.5091 \\
Qwen2.5 & AG News & 8 & 0.5384 & 0.5442 & 0.5442 & 0.5361 & 0.5183 \\
Qwen2.5 & AG News & 32 & 0.6344 & 0.6472 & 0.6472 & 0.6293 & 0.5261 \\
Qwen2.5 & IMDb & 8 & 0.5070 & 0.5084 & 0.5084 & 0.5143 & 0.5011 \\
Qwen2.5 & IMDb & 32 & 0.5236 & 0.5212 & 0.5212 & 0.5333 & 0.5083 \\
Qwen3 & Emotion & 8 & 0.5139 & 0.5135 & 0.5135 & 0.5147 & 0.5048 \\
Qwen3 & Emotion & 32 & 0.5270 & 0.5275 & 0.5275 & 0.5300 & 0.5089 \\
Qwen3 & AG News & 8 & 0.5499 & 0.5478 & 0.5478 & 0.5477 & 0.5160 \\
Qwen3 & AG News & 32 & 0.6508 & 0.6468 & 0.6468 & 0.6478 & 0.5248 \\
Qwen3 & IMDb & 8 & 0.5100 & 0.5014 & 0.5014 & 0.5022 & 0.5066 \\
Qwen3 & IMDb & 32 & 0.5343 & 0.5225 & 0.5225 & 0.5132 & 0.5048 \\
SmolLM2 & Emotion & 8 & 0.5077 & 0.5052 & 0.5052 & 0.5091 & 0.5061 \\
SmolLM2 & Emotion & 32 & 0.5247 & 0.5086 & 0.5086 & 0.5241 & 0.5108 \\
SmolLM2 & AG News & 8 & 0.5518 & 0.5524 & 0.5524 & 0.5476 & 0.5138 \\
SmolLM2 & AG News & 32 & 0.6589 & 0.6461 & 0.6461 & 0.6484 & 0.5222 \\
SmolLM2 & IMDb & 8 & 0.5058 & 0.5031 & 0.5031 & 0.5044 & 0.5050 \\
SmolLM2 & IMDb & 32 & 0.5181 & 0.5123 & 0.5123 & 0.5333 & 0.5088 \\
\bottomrule
\end{tabular}

\end{table}

\begin{table}[!ht]
  \centering
  \caption{Raw-coordinate macro-AUROC with the same feature-selection and scoring procedure.
  These probes do not have an SAE seed or a Top-$K$ setting. Logits, RC-DS, and Std. tie
  because positive per-coordinate scaling and constant shifts preserve rankings and AUROC.}
  \label{tab:full-semantic-raw}
  \begin{tabular}{llrrrrrr}
\toprule
Model & Task & $K$ & Logits & RC-DS & Hidden & Std. & ZCA \\
\midrule
Qwen2.5 & Emotion & -- & 0.5669 & 0.5669 & 0.5815 & 0.5669 & 0.5268 \\
Qwen2.5 & AG News & -- & 0.8493 & 0.8493 & 0.8580 & 0.8493 & 0.5324 \\
Qwen2.5 & IMDb & -- & 0.5650 & 0.5650 & 0.6886 & 0.5650 & 0.5354 \\
Qwen3 & Emotion & -- & 0.5958 & 0.5958 & 0.5841 & 0.5958 & 0.5165 \\
Qwen3 & AG News & -- & 0.8471 & 0.8471 & 0.8664 & 0.8471 & 0.5449 \\
Qwen3 & IMDb & -- & 0.5564 & 0.5564 & 0.6233 & 0.5564 & 0.5094 \\
SmolLM2 & Emotion & -- & 0.5650 & 0.5650 & 0.5713 & 0.5650 & 0.5005 \\
SmolLM2 & AG News & -- & 0.8716 & 0.8716 & 0.8313 & 0.8716 & 0.5514 \\
SmolLM2 & IMDb & -- & 0.6241 & 0.6241 & 0.5797 & 0.6241 & 0.5563 \\
\bottomrule
\end{tabular}

\end{table}

\subsection{Reconstruction and training curves}
\label{sec:paired-reconstruction}

\begin{figure}[!ht]
  \centering
  \includegraphics[width=\linewidth]{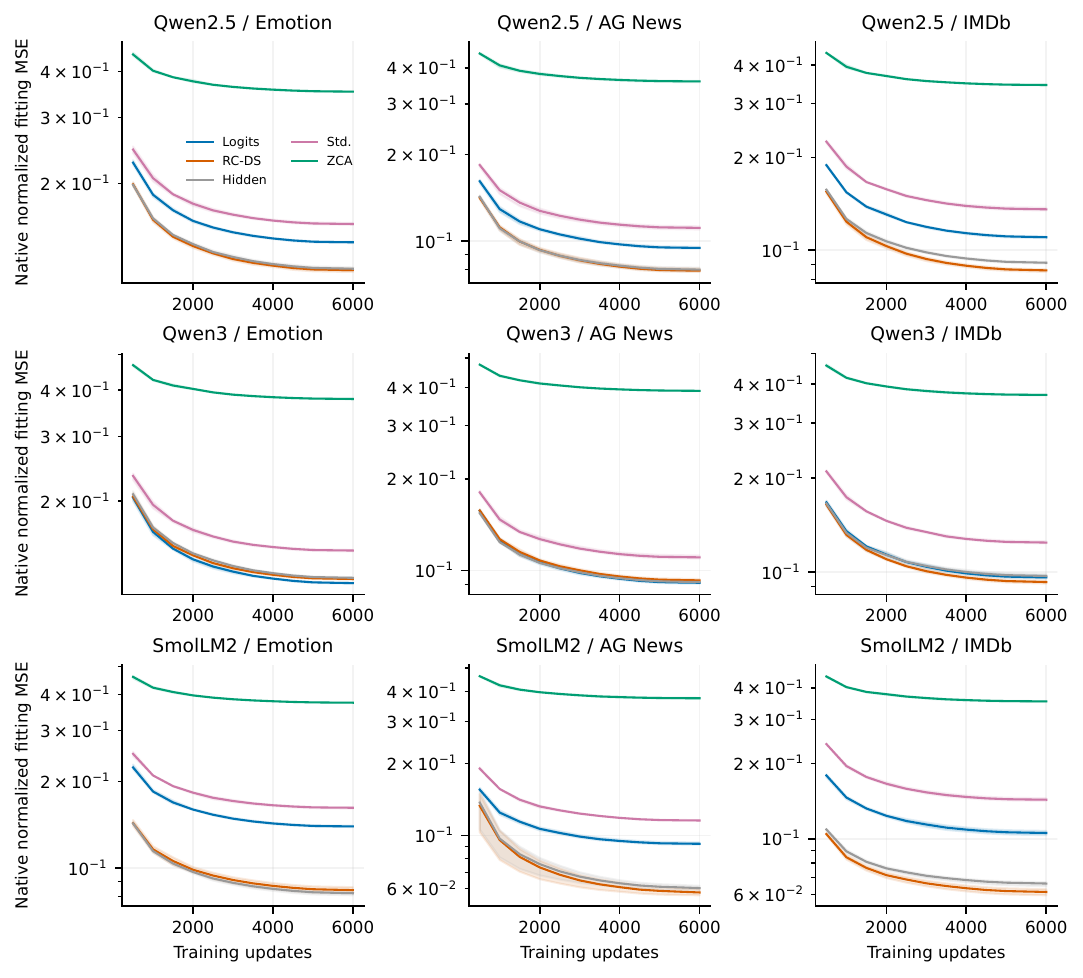}
  \caption{Fitting curves for $K=8$. Lines average three selected learning-rate fits, one per
  SAE seed. Shading spans the seeds, not a confidence interval. Loss is native normalized
  fitting MSE, with a logarithmic vertical axis. Qwen2.5 is development data.}
  \label{fig:full-training-eight}
\end{figure}

\begin{figure}[!ht]
  \centering
  \includegraphics[width=\linewidth]{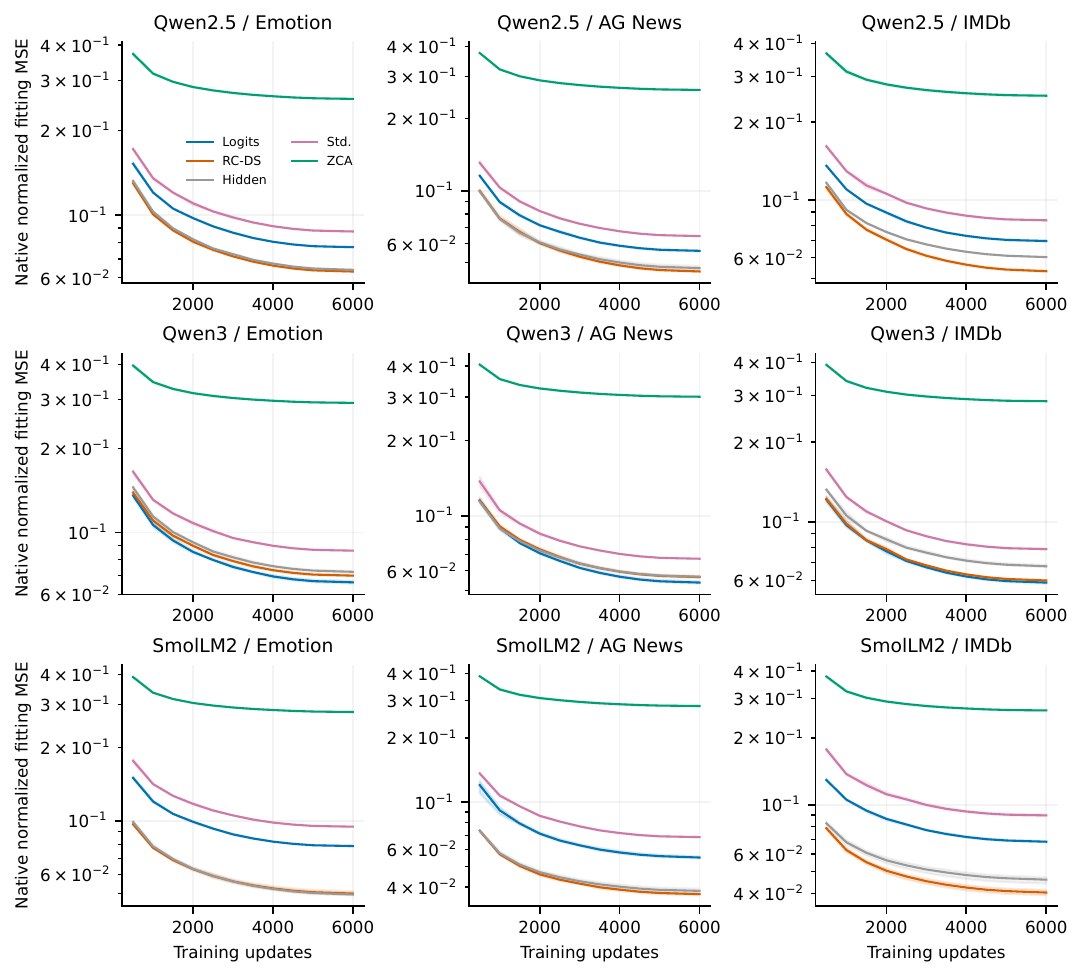}
  \caption{Fitting curves for $K=32$, using the same conventions as
  Figure~\ref{fig:full-training-eight}. Both candidate learning rates and all histories remain
  in the internal study records, including rates not selected for scoring.}
  \label{fig:full-training-thirtytwo}
\end{figure}

Tables~\ref{tab:full-semantic-logit-ev} and~\ref{tab:full-semantic-gainfree-ev} score the same
selected models on the disjoint scoring texts, using common logit and gain-free metrics.
These are different questions from detecting semantic labels. The fitting curves in
Figures~\ref{fig:full-training-eight} and~\ref{fig:full-training-thirtytwo} instead use each
arm's native RMS-normalized reconstruction loss. Their vertical values should not be treated
as a common-metric ranking of coordinate methods. A fixed training budget and declining loss
do not prove convergence.

We report paired differences separately for each checkpoint, task, and sparsity level.
For metric $E$ and baseline $b$, a seed-level difference is
$d_s=E(\mathrm{RC\text{-}DS},s)-E(b,s)$, using the originally selected fits for the same seed.
Tables~\ref{tab:paired-logits}--\ref{tab:paired-zca} give the mean and observed range of the
three differences. Ranges describe run-to-run variation on fixed texts and checkpoints.
They are not confidence intervals and do not measure uncertainty from resampling texts or
choosing other models. All comparisons are retained regardless of size or sign. A range
that spans zero shows a reversal among the observed seeds. A range entirely on one side
of zero is still not a population-level significance claim.

For Table~\ref{tab:reconstruction-summary}, we first average the six task--sparsity
differences within each checkpoint and seed, then report the mean and sample standard
deviation of those three averages. This preserves seed matching instead of pooling correlated
settings as independent observations. We do not fit a population random-effects model to
three deliberately selected checkpoints. The setting counts describe this grid only.

\begin{table}[!ht]
  \centering\small
  \caption{Paired RC-DS minus centered-logit reconstruction differences. Each metric gives
  the mean and [minimum, maximum] across seeds 730201--730203, not a confidence interval.
  Positive values favor RC-DS. Tasks and sparsity levels repeat within each fixed checkpoint.}
  \label{tab:paired-logits}
  \begin{tabular}{llrrr}
\toprule
Model & Task & $K$ & Gain-free EV & Logit EV \\
\midrule
Qwen2.5 & Emotion & 8 & $+0.0353\ [+0.0318,+0.0398]$ & $-0.0041\ [-0.0069,-0.0025]$ \\
Qwen2.5 & Emotion & 32 & $+0.0305\ [+0.0265,+0.0349]$ & $-0.0055\ [-0.0083,-0.0027]$ \\
Qwen2.5 & AG News & 8 & $+0.0108\ [+0.0092,+0.0131]$ & $-0.0046\ [-0.0058,-0.0039]$ \\
Qwen2.5 & AG News & 32 & $+0.0114\ [+0.0104,+0.0122]$ & $-0.0067\ [-0.0072,-0.0065]$ \\
Qwen2.5 & IMDb & 8 & $+0.0242\ [+0.0216,+0.0261]$ & $-0.0036\ [-0.0042,-0.0032]$ \\
Qwen2.5 & IMDb & 32 & $+0.0177\ [+0.0160,+0.0189]$ & $-0.0101\ [-0.0106,-0.0095]$ \\
Qwen3 & Emotion & 8 & $+0.0134\ [+0.0079,+0.0215]$ & $+0.0006\ [-0.0056,+0.0086]$ \\
Qwen3 & Emotion & 32 & $+0.0080\ [+0.0065,+0.0099]$ & $-0.0080\ [-0.0096,-0.0056]$ \\
Qwen3 & AG News & 8 & $+0.0048\ [+0.0031,+0.0072]$ & $-0.0023\ [-0.0046,+0.0008]$ \\
Qwen3 & AG News & 32 & $+0.0069\ [+0.0052,+0.0099]$ & $-0.0032\ [-0.0049,-0.0002]$ \\
Qwen3 & IMDb & 8 & $+0.0136\ [+0.0101,+0.0177]$ & $-0.0003\ [-0.0034,+0.0027]$ \\
Qwen3 & IMDb & 32 & $+0.0075\ [+0.0073,+0.0077]$ & $-0.0065\ [-0.0068,-0.0061]$ \\
SmolLM2 & Emotion & 8 & $+0.0968\ [+0.0894,+0.1044]$ & $-0.0188\ [-0.0200,-0.0182]$ \\
SmolLM2 & Emotion & 32 & $+0.0553\ [+0.0525,+0.0585]$ & $-0.0371\ [-0.0424,-0.0334]$ \\
SmolLM2 & AG News & 8 & $+0.0460\ [+0.0437,+0.0472]$ & $-0.0309\ [-0.0351,-0.0276]$ \\
SmolLM2 & AG News & 32 & $+0.0318\ [+0.0287,+0.0367]$ & $-0.0354\ [-0.0390,-0.0296]$ \\
SmolLM2 & IMDb & 8 & $+0.0703\ [+0.0635,+0.0747]$ & $-0.0127\ [-0.0148,-0.0107]$ \\
SmolLM2 & IMDb & 32 & $+0.0391\ [+0.0374,+0.0400]$ & $-0.0281\ [-0.0285,-0.0279]$ \\
\bottomrule
\end{tabular}

\end{table}

\begin{table}[!ht]
  \centering\small
  \caption{Paired RC-DS minus hidden-state reconstruction differences, with the same
  mean and observed seed-range convention as Table~\ref{tab:paired-logits}.}
  \label{tab:paired-hidden}
  \begin{tabular}{llrrr}
\toprule
Model & Task & $K$ & Gain-free EV & Logit EV \\
\midrule
Qwen2.5 & Emotion & 8 & $+0.0013\ [-0.0003,+0.0027]$ & $+0.0018\ [-0.0017,+0.0041]$ \\
Qwen2.5 & Emotion & 32 & $+0.0048\ [+0.0003,+0.0098]$ & $+0.0054\ [+0.0006,+0.0100]$ \\
Qwen2.5 & AG News & 8 & $+0.0064\ [+0.0050,+0.0078]$ & $+0.0081\ [+0.0074,+0.0089]$ \\
Qwen2.5 & AG News & 32 & $+0.0080\ [+0.0062,+0.0090]$ & $+0.0095\ [+0.0067,+0.0113]$ \\
Qwen2.5 & IMDb & 8 & $+0.0004\ [-0.0012,+0.0027]$ & $+0.0006\ [-0.0001,+0.0013]$ \\
Qwen2.5 & IMDb & 32 & $+0.0050\ [+0.0011,+0.0075]$ & $+0.0061\ [+0.0022,+0.0083]$ \\
Qwen3 & Emotion & 8 & $+0.0081\ [+0.0038,+0.0115]$ & $+0.0068\ [+0.0016,+0.0104]$ \\
Qwen3 & Emotion & 32 & $+0.0103\ [+0.0049,+0.0146]$ & $+0.0098\ [+0.0045,+0.0142]$ \\
Qwen3 & AG News & 8 & $+0.0032\ [+0.0025,+0.0045]$ & $+0.0034\ [+0.0024,+0.0048]$ \\
Qwen3 & AG News & 32 & $+0.0060\ [+0.0048,+0.0084]$ & $+0.0057\ [+0.0041,+0.0078]$ \\
Qwen3 & IMDb & 8 & $+0.0039\ [+0.0028,+0.0048]$ & $+0.0031\ [+0.0015,+0.0043]$ \\
Qwen3 & IMDb & 32 & $+0.0124\ [+0.0111,+0.0132]$ & $+0.0122\ [+0.0109,+0.0130]$ \\
SmolLM2 & Emotion & 8 & $+0.0005\ [-0.0006,+0.0018]$ & $+0.0040\ [+0.0017,+0.0055]$ \\
SmolLM2 & Emotion & 32 & $+0.0032\ [+0.0003,+0.0061]$ & $+0.0045\ [+0.0002,+0.0073]$ \\
SmolLM2 & AG News & 8 & $+0.0003\ [-0.0022,+0.0042]$ & $+0.0063\ [+0.0014,+0.0115]$ \\
SmolLM2 & AG News & 32 & $+0.0083\ [+0.0065,+0.0099]$ & $+0.0176\ [+0.0135,+0.0249]$ \\
SmolLM2 & IMDb & 8 & $+0.0002\ [-0.0008,+0.0018]$ & $+0.0138\ [+0.0118,+0.0175]$ \\
SmolLM2 & IMDb & 32 & $+0.0006\ [-0.0003,+0.0023]$ & $+0.0079\ [+0.0060,+0.0098]$ \\
\bottomrule
\end{tabular}

\end{table}

\begin{table}[!ht]
  \centering\small
  \caption{Paired RC-DS minus standardized-coordinate reconstruction differences.
  Entries give the mean and observed range over three matched seeds.}
  \label{tab:paired-standardized}
  \begin{tabular}{llrrr}
\toprule
Model & Task & $K$ & Gain-free EV & Logit EV \\
\midrule
Qwen2.5 & Emotion & 8 & $+0.0644\ [+0.0582,+0.0726]$ & $+0.0127\ [+0.0095,+0.0154]$ \\
Qwen2.5 & Emotion & 32 & $+0.0613\ [+0.0572,+0.0686]$ & $+0.0143\ [+0.0108,+0.0200]$ \\
Qwen2.5 & AG News & 8 & $+0.0252\ [+0.0235,+0.0280]$ & $-0.0007\ [-0.0017,+0.0001]$ \\
Qwen2.5 & AG News & 32 & $+0.0277\ [+0.0258,+0.0288]$ & $+0.0022\ [+0.0009,+0.0036]$ \\
Qwen2.5 & IMDb & 8 & $+0.0777\ [+0.0685,+0.0865]$ & $+0.0197\ [+0.0166,+0.0240]$ \\
Qwen2.5 & IMDb & 32 & $+0.0496\ [+0.0478,+0.0508]$ & $+0.0093\ [+0.0084,+0.0105]$ \\
Qwen3 & Emotion & 8 & $+0.0332\ [+0.0281,+0.0372]$ & $+0.0304\ [+0.0265,+0.0333]$ \\
Qwen3 & Emotion & 32 & $+0.0352\ [+0.0320,+0.0392]$ & $+0.0348\ [+0.0314,+0.0389]$ \\
Qwen3 & AG News & 8 & $+0.0122\ [+0.0103,+0.0150]$ & $+0.0101\ [+0.0082,+0.0127]$ \\
Qwen3 & AG News & 32 & $+0.0162\ [+0.0147,+0.0175]$ & $+0.0158\ [+0.0142,+0.0167]$ \\
Qwen3 & IMDb & 8 & $+0.0388\ [+0.0364,+0.0411]$ & $+0.0228\ [+0.0200,+0.0254]$ \\
Qwen3 & IMDb & 32 & $+0.0277\ [+0.0261,+0.0289]$ & $+0.0227\ [+0.0209,+0.0240]$ \\
SmolLM2 & Emotion & 8 & $+0.1626\ [+0.1564,+0.1704]$ & $+0.0077\ [+0.0054,+0.0114]$ \\
SmolLM2 & Emotion & 32 & $+0.1276\ [+0.1227,+0.1305]$ & $-0.0059\ [-0.0104,-0.0028]$ \\
SmolLM2 & AG News & 8 & $+0.0822\ [+0.0740,+0.0908]$ & $-0.0142\ [-0.0226,-0.0085]$ \\
SmolLM2 & AG News & 32 & $+0.0700\ [+0.0685,+0.0710]$ & $-0.0132\ [-0.0142,-0.0125]$ \\
SmolLM2 & IMDb & 8 & $+0.1715\ [+0.1669,+0.1795]$ & $+0.0265\ [+0.0247,+0.0288]$ \\
SmolLM2 & IMDb & 32 & $+0.1107\ [+0.1055,+0.1176]$ & $+0.0066\ [+0.0048,+0.0090]$ \\
\bottomrule
\end{tabular}

\end{table}

\begin{table}[!ht]
  \centering\small
  \caption{Paired RC-DS minus ZCA reconstruction differences.
  Entries give the mean and observed range over three matched seeds.}
  \label{tab:paired-zca}
  \begin{tabular}{llrrr}
\toprule
Model & Task & $K$ & Gain-free EV & Logit EV \\
\midrule
Qwen2.5 & Emotion & 8 & $+0.3025\ [+0.2962,+0.3110]$ & $+0.2570\ [+0.2559,+0.2589]$ \\
Qwen2.5 & Emotion & 32 & $+0.4759\ [+0.4687,+0.4796]$ & $+0.4301\ [+0.4240,+0.4366]$ \\
Qwen2.5 & AG News & 8 & $+0.3112\ [+0.3029,+0.3238]$ & $+0.2811\ [+0.2720,+0.2915]$ \\
Qwen2.5 & AG News & 32 & $+0.3696\ [+0.3651,+0.3777]$ & $+0.3429\ [+0.3350,+0.3483]$ \\
Qwen2.5 & IMDb & 8 & $+0.3131\ [+0.2944,+0.3226]$ & $+0.2670\ [+0.2462,+0.2779]$ \\
Qwen2.5 & IMDb & 32 & $+0.4231\ [+0.4179,+0.4261]$ & $+0.3855\ [+0.3806,+0.3901]$ \\
Qwen3 & Emotion & 8 & $+0.3617\ [+0.3527,+0.3673]$ & $+0.3610\ [+0.3520,+0.3664]$ \\
Qwen3 & Emotion & 32 & $+0.5368\ [+0.5208,+0.5508]$ & $+0.5412\ [+0.5276,+0.5546]$ \\
Qwen3 & AG News & 8 & $+0.3488\ [+0.3462,+0.3523]$ & $+0.3474\ [+0.3446,+0.3508]$ \\
Qwen3 & AG News & 32 & $+0.4175\ [+0.4162,+0.4189]$ & $+0.4235\ [+0.4198,+0.4264]$ \\
Qwen3 & IMDb & 8 & $+0.3498\ [+0.3475,+0.3533]$ & $+0.3411\ [+0.3385,+0.3450]$ \\
Qwen3 & IMDb & 32 & $+0.4467\ [+0.4389,+0.4515]$ & $+0.4473\ [+0.4404,+0.4538]$ \\
SmolLM2 & Emotion & 8 & $+0.4238\ [+0.4092,+0.4373]$ & $+0.2495\ [+0.2389,+0.2552]$ \\
SmolLM2 & Emotion & 32 & $+0.5273\ [+0.5213,+0.5322]$ & $+0.3874\ [+0.3813,+0.3930]$ \\
SmolLM2 & AG News & 8 & $+0.4010\ [+0.3979,+0.4068]$ & $+0.2892\ [+0.2864,+0.2930]$ \\
SmolLM2 & AG News & 32 & $+0.4286\ [+0.4182,+0.4442]$ & $+0.3767\ [+0.3725,+0.3822]$ \\
SmolLM2 & IMDb & 8 & $+0.4462\ [+0.4315,+0.4568]$ & $+0.3269\ [+0.3128,+0.3410]$ \\
SmolLM2 & IMDb & 32 & $+0.4736\ [+0.4703,+0.4799]$ & $+0.4075\ [+0.3985,+0.4121]$ \\
\bottomrule
\end{tabular}

\end{table}

\FloatBarrier

\begin{table}[!ht]
  \centering
  \caption{Scoring-set explained variance in the same full centered-logit metric, averaging
  three selected SAE seeds. Higher is better.}
  \label{tab:full-semantic-logit-ev}
  \begin{tabular}{llrrrrrr}
\toprule
Model & Task & $K$ & Logits & RC-DS & Hidden & Std. & ZCA \\
\midrule
Qwen2.5 & Emotion & 8 & 0.3930 & 0.3889 & 0.3870 & 0.3762 & 0.1319 \\
Qwen2.5 & Emotion & 32 & 0.5207 & 0.5152 & 0.5098 & 0.5008 & 0.0851 \\
Qwen2.5 & AG News & 8 & 0.6325 & 0.6279 & 0.6198 & 0.6285 & 0.3467 \\
Qwen2.5 & AG News & 32 & 0.7002 & 0.6935 & 0.6840 & 0.6912 & 0.3506 \\
Qwen2.5 & IMDb & 8 & 0.5403 & 0.5367 & 0.5361 & 0.5170 & 0.2697 \\
Qwen2.5 & IMDb & 32 & 0.6167 & 0.6066 & 0.6006 & 0.5973 & 0.2212 \\
Qwen3 & Emotion & 8 & 0.4663 & 0.4668 & 0.4601 & 0.4364 & 0.1058 \\
Qwen3 & Emotion & 32 & 0.6040 & 0.5960 & 0.5862 & 0.5612 & 0.0549 \\
Qwen3 & AG News & 8 & 0.6451 & 0.6428 & 0.6393 & 0.6326 & 0.2953 \\
Qwen3 & AG News & 32 & 0.7223 & 0.7191 & 0.7134 & 0.7033 & 0.2956 \\
Qwen3 & IMDb & 8 & 0.6059 & 0.6056 & 0.6025 & 0.5828 & 0.2645 \\
Qwen3 & IMDb & 32 & 0.6835 & 0.6770 & 0.6649 & 0.6543 & 0.2297 \\
SmolLM2 & Emotion & 8 & 0.4125 & 0.3937 & 0.3897 & 0.3860 & 0.1441 \\
SmolLM2 & Emotion & 32 & 0.5422 & 0.5051 & 0.5006 & 0.5109 & 0.1177 \\
SmolLM2 & AG News & 8 & 0.6410 & 0.6101 & 0.6038 & 0.6243 & 0.3209 \\
SmolLM2 & AG News & 32 & 0.7153 & 0.6799 & 0.6623 & 0.6931 & 0.3032 \\
SmolLM2 & IMDb & 8 & 0.5748 & 0.5621 & 0.5483 & 0.5356 & 0.2351 \\
SmolLM2 & IMDb & 32 & 0.6471 & 0.6189 & 0.6111 & 0.6123 & 0.2114 \\
\bottomrule
\end{tabular}

\end{table}

\begin{table}[!ht]
  \centering
  \caption{Scoring-set explained variance after removing readout gains, averaging three
  selected SAE seeds. Higher is better. This weights errors differently from logit EV.
  Entries are rounded to four decimal places. Differences reported in the text are
  computed from unrounded means, then rounded separately.}
  \label{tab:full-semantic-gainfree-ev}
  \begin{tabular}{llrrrrrr}
\toprule
Model & Task & $K$ & Logits & RC-DS & Hidden & Std. & ZCA \\
\midrule
Qwen2.5 & Emotion & 8 & 0.4269 & 0.4622 & 0.4609 & 0.3977 & 0.1596 \\
Qwen2.5 & Emotion & 32 & 0.5700 & 0.6006 & 0.5958 & 0.5392 & 0.1247 \\
Qwen2.5 & AG News & 8 & 0.6673 & 0.6781 & 0.6717 & 0.6529 & 0.3668 \\
Qwen2.5 & AG News & 32 & 0.7383 & 0.7497 & 0.7416 & 0.7220 & 0.3800 \\
Qwen2.5 & IMDb & 8 & 0.6086 & 0.6328 & 0.6324 & 0.5551 & 0.3196 \\
Qwen2.5 & IMDb & 32 & 0.6916 & 0.7092 & 0.7042 & 0.6596 & 0.2861 \\
Qwen3 & Emotion & 8 & 0.4576 & 0.4710 & 0.4629 & 0.4378 & 0.1093 \\
Qwen3 & Emotion & 32 & 0.5884 & 0.5964 & 0.5861 & 0.5611 & 0.0596 \\
Qwen3 & AG News & 8 & 0.6472 & 0.6520 & 0.6487 & 0.6398 & 0.3031 \\
Qwen3 & AG News & 32 & 0.7193 & 0.7262 & 0.7201 & 0.7099 & 0.3086 \\
Qwen3 & IMDb & 8 & 0.6095 & 0.6231 & 0.6192 & 0.5843 & 0.2733 \\
Qwen3 & IMDb & 32 & 0.6846 & 0.6921 & 0.6797 & 0.6645 & 0.2454 \\
SmolLM2 & Emotion & 8 & 0.5428 & 0.6397 & 0.6392 & 0.4771 & 0.2159 \\
SmolLM2 & Emotion & 32 & 0.6951 & 0.7504 & 0.7472 & 0.6228 & 0.2231 \\
SmolLM2 & AG News & 8 & 0.7520 & 0.7980 & 0.7977 & 0.7158 & 0.3970 \\
SmolLM2 & AG News & 32 & 0.8267 & 0.8585 & 0.8502 & 0.7885 & 0.4299 \\
SmolLM2 & IMDb & 8 & 0.7286 & 0.7988 & 0.7987 & 0.6273 & 0.3526 \\
SmolLM2 & IMDb & 32 & 0.8095 & 0.8486 & 0.8480 & 0.7379 & 0.3750 \\
\bottomrule
\end{tabular}

\end{table}

\subsection{Provenance and internal verification}

The internally retained protocol pins all model and dataset revisions. Its SHA-256 begins
\texttt{7fcfe2e0d79652b1}. The source code is also locked, separately from its outputs.
Saved forecasts precede all 360 reserved fits. Checkpoint replay verifies 1,080 training
checkpoints across 540
fits and 585 semantic score rows, including trained, untrained, and raw-coordinate controls.
It recomputes feature selection and exact tied-score AUROC and checks text/token separation,
selected learning rates, source hashes, and forecast timestamps.

SAE training uses three concurrent CPU workers with one thread per fit. Development training
takes 2.66 wall-clock hours and confirmation 5.62 hours on the recorded machine, excluding
representation extraction and final scoring. Summed per-fit training times are 7.37 and
14.08 hours. These measurements are not portable runtime guarantees.

Development and confirmation outputs come from separate executions of identical
checksum-locked source. Their metadata records the phase and source hash. An initial setup
failure occurred before SAE training and remains in the internal record. Tables and figures
are generated from saved results without retraining. The source code and executable notebooks
are not publicly released.

The scope remains three fixed checkpoints, with only two reserved for confirmation.
SAE seeds and tasks are repeated measurements, not additional independently trained models.
The outcome is a single-feature label-detection score on the specified text partitions.
It is not a test of all possible semantics, latent-factor truth, or causal intervention quality.

\end{document}